# Lightweight high-speed and high-force gripper for assembly

Toshihiro Nishimura, *Member, IEEE*, Takeshi Takaki, *Member, IEEE*, Yosuke Suzuki, *Member, IEEE*, Tokuo Tsuji, *Member, IEEE*, and Tetsuyou Watanabe, *Member, IEEE*

***Abstract*—This paper presents a novel industrial robotic gripper with a high grasping speed (maximum: 1396 mm/s), high tip force (maximum: 80 N) for grasping, large motion range, and lightweight design (0.3 kg). To realize these features, the high-speed section of the quick-return mechanism and load-sensitive continuously variable transmission mechanism are installed in the gripper. The gripper is also equipped with a self-centering function. The high grasping speed and self-centering function improve the cycle time in robotic operations. In addition, the high tip force is advantageous for stably grasping and assembling heavy objects. Moreover, the design of the gripper reduce the gripper's proportion of the manipulator's payload, thus increasing the weight of the object that can be grasped. The gripper performance was validated through kinematic and static analyses as well as experimental evaluations. This paper also presents the analysis of the self-centering function of the developed gripper.**

## I. INTRODUCTION

THIS paper proposes a novel lightweight robotic gripper for realizing a high closing/opening speed and high tip force. Although several robotic grippers have been developed for factory automation [1], developing systems with high production capacities remains challenging. To achieve this, minimization of the cycle time through high-speed motion is preferred for robotic hands. However, the motion speed and tip force, which are preferred for stably grasping objects while counteracting disturbing forces from any direction in assembly tasks and transferring heavy objects, have a tradeoff relationship. A large actuator attains both high speed and force but is often unsuitable for applications because it requires a correspondingly large and heavy gripper. This study addresses the challenge of developing a lightweight robotic gripper that realizes both high-speed closing/opening motion of the fingers and high tip force using a single low-power actuator. Several technologies have been developed to realize robotic grippers with high speeds or forces. In [2][3][4][5], although mechanisms to amplify the tip force were installed, finger speed was not considered. Several researchers have focused on the motion speed of robotic hands [6]–[8]. These studies used a low-reduction-ratio mechanism to achieve a high-speed motion; thus, a large tip force could not be obtained. Several studies developed robotic hands to realize high-speed and high-force grasping [9][10]. They realized these two features by switching the operation modes, i.e., the high-speed and high-force modes, using two or more motors. An increase in the number of installed motors increased the weight of the grippers. Although robotic hands with high speed and force exist, no attempt has been made to realize a robotic gripper that realizes them using a single motor for a lightweight design.

The developed gripper is shown in Fig. 1. To realize high-speed motion and a high tip force using a lightweight design, the developed gripper employs two key mechanisms: a quick-return mechanism and load-sensitive continuously variable transmission (LS-CVT) [2]. The high-speed section of the quick-return mechanism is extracted and adopted to obtain the high-speed motion for approaching and releasing the target object. This mechanism extended the range of movement of the fingertips, enabling the gripper to grasp objects of various sizes. The tip force is increased by the LS-CVT after the fingertip comes into contact with the target object. The fingers of the developed gripper are driven by a single-input shaft to which the LS-CVT is installed. This structure reduces the number of LS-CVTs (installed on each finger in [2]), thereby enabling a lightweight design. The integration of the LS-CVT and quick-return mechanism provides not only high-speed and high-force grasping but also the following benefits: 1) an impulsive force applied to the fingers is absorbed by the LS-CVT owing to its high back-drivability; 2) the LS-CVT can amplify not only the tip force but also the opening/closing speed if the design parameters are set appropriately; and 3) the motion range of the LS-CVT is limited because of its small design and link-mechanism-based structure, which is remarkably extended by the high-speed section of the quick-return mechanism. In the developed gripper, a radial arrangement of fingers is employed to counteract the disturbing forces in any direction. The arrangement and synchronized finger motion further enables the gripper to align the position and posture of the target objects, i.e., a self-centering function is available. As proposed in several studies [6][5][11][12][13], a robotic hand with a self-centering function is effective for accomplishing tasks that require precise object positioning such as assembly.

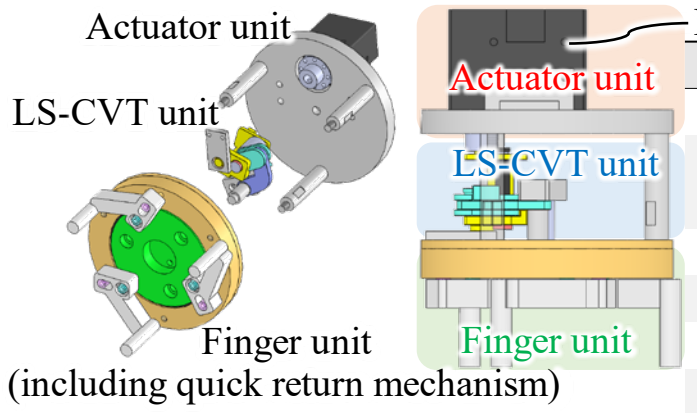

Motor (Dynamixel, XM430-W210)

| Specification and performance | |
|---|---|
| Set torque in this paper | 1300 Nmm |
| Maximum velocity | 412 °/s (68.7 rpm) |
| Weight | 82 g |
| Diameter | 85 mm |
| Height | 121 mm |
| Control method | Current |

**Fig. 1.** Gripper developed in this study

T. Nishimura, Y. Suzuki, T. Tsuji, and T. Watanabe are with the Faculty of Frontier Engineering, Institute of Science and Engineering, Kanazawa University, Kakuma-machi, Kanazawa city, Ishikawa, 9201192 Japan (e-mail: tnishimura@se.kanazawa-u.ac.jp, te-watanabe@ieee.org).

T. Takaki is with the Graduate School of Advanced Science and Engineering, Hiroshima University, 1-4-1 Kagamiyama, Higashi-Hiroshima City, Hiroshima, 7398527 Japan

## II. Gripper Design

### *A. Functional requirements*

To render the gripper suitable for industrial applications, such as product assembly, the following design requirements were set: 1) grasping motion speed, i.e., opening/closing speed of the finger, exceeding 1200 mm/s; 2) tip force exceeding 50 N; 3) gripper mass of less than 0.3 kg; 4) capability of grasping objects with a thinness of 1 mm or more; 5) width of graspable objects exceeding 70 mm; 6) self-centering function; and 7) main targets are radially symmetric cylindrical parts. The payload, mass, and object size requirements were set assuming that the gripper was installed in small industrial robots. The requirements for graspable objects were determined based on objects used in the World Robot Summit (WRS) Assembly Challenge [14].

### *B. Structure and Main mechanisms*

Fig. 2 details the structure of the developed gripper shown in Fig. 1. In the actuator unit, a servomotor is supported by the base of the finger unit (orange part in Figs. 1 and 2), and the output shaft of the motor is connected to the LS-CVT unit. The fingers are driven by the rotational motion of a rotational plate (green part), which in turn is impelled by the motor rotation through the LS-CVT unit.

*1) Quick-return mechanism*

To achieve a high-speed grasping motion, a quick-return mechanism was introduced, as illustrated in Fig. 3. The opening/closing mechanism of the developed gripper was designed such that the fingers were driven using only the high-speed section of the quick-return mechanism. The mechanism consists of three fingers: a base, a rotational plate, three inner pins, and three outer pins. Each finger is attached to the rotational plate and base via inner and outer pins; that is, the inner and outer pins are fixed to the rotational plate and base, respectively. The plate is rotated relative to the base using a motor. Correspondingly, the inner pin moves relative to the outer pin to close or open the fingers. Considering the high-speed section of the quick-return mechanism, these factors are designed to obtain a large movement range for the fingertip with a small input movement range. Furthermore, the three fingers were arranged at equal intervals in a radial pattern such that the simultaneous closing motion toward the center provided a self-centering function.

*2) LS-CVT*

To obtain a high force irrespective of the increasing high-speed ratio of the quick-return mechanism, the LS-CVT [2] was adopted. As shown in Fig. 4, the LS-CVT features a five-link mechanism, in which the input link is separated via a torsion spring, and the other links are the same as those of a conventional four-link mechanism. Let $P_i$ ($i \in \{1 \cdots 5\}$) be the $i$-th joint, as shown in Fig. 4(a). Because the two separated input links 1 and 2 work together as joint input link owing to the torsion spring, a virtual input link between joints $P_2$ and $P_4$ is assumed. Let $l_{in_v}$ be the length $\overline{P_2P_4}$, $l_{out}$ be the length of the output link, $\theta_1$ be the angle between the input link and the normal line from joint $P_2$ to the floating link, $\theta_2$ be the angle between the output link and the normal line from joint $P_1$ to the floating link, and $\tau_{in}$ and $\omega_{in}$ be the motor input torque and angular velocity, respectively. $\tau_{out}$ and $\omega_{out}$ are given by:

$$\tau_{out} = \frac{l_{out}\cos\theta_2}{l_{in_v}\cos\theta_1}\tau_{in} \quad (1)$$

$$\omega_{out} = \frac{l_{in_v}\cos\theta_1}{l_{out}\cos\theta_2}\omega_{in} \quad (2)$$

Note that the torques from the torsion spring applied to the separated input links 1 and 2 cancel each other internally; thus, the torques do not appear in (1) and (2).

When an external load is not applied to the output link, the separated input links 1 and 2 behave as one link. An external

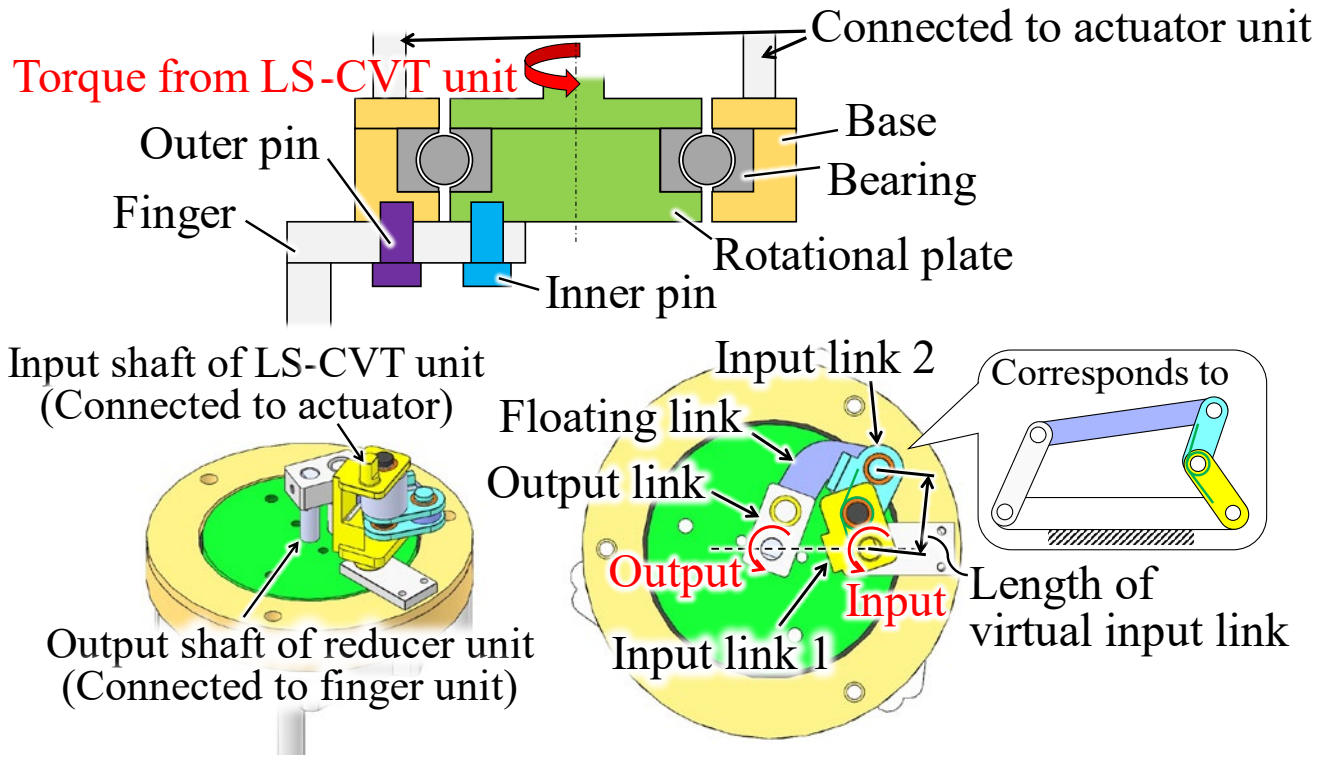

**Fig. 2.** Structure of the developed gripper

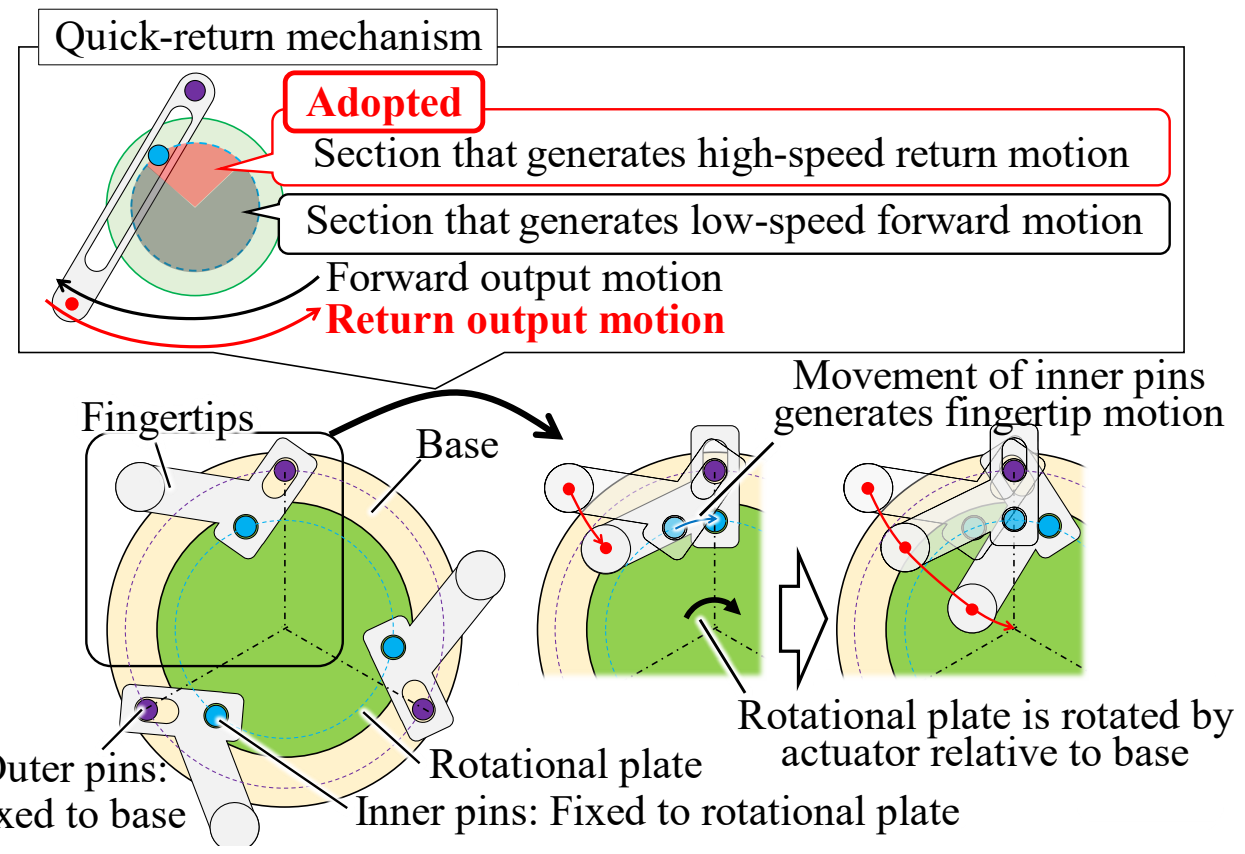

**Fig. 3.** Opening/closing mechanism of the developed gripper using high-speed section of quick-return mechanism

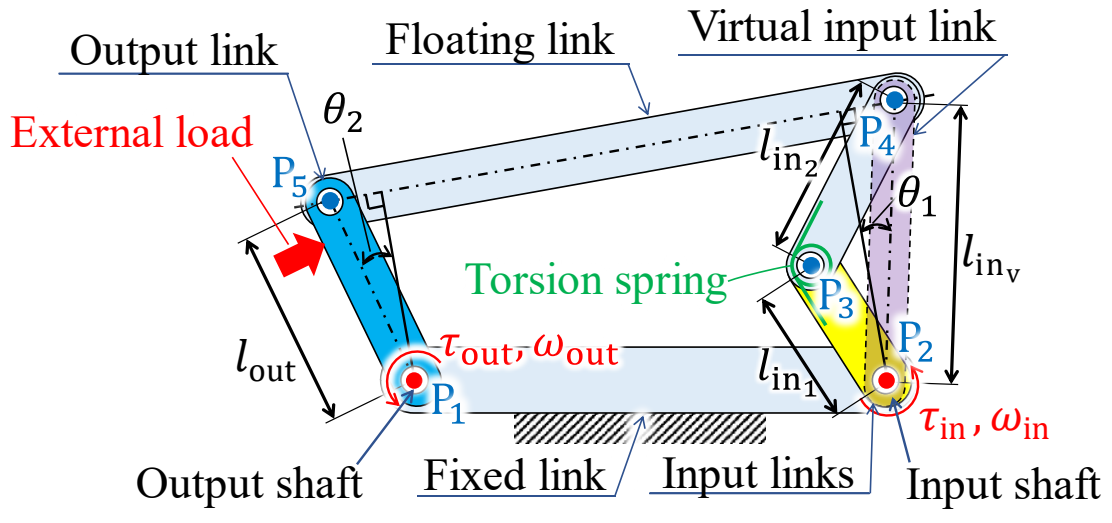

(a) Mechanical model

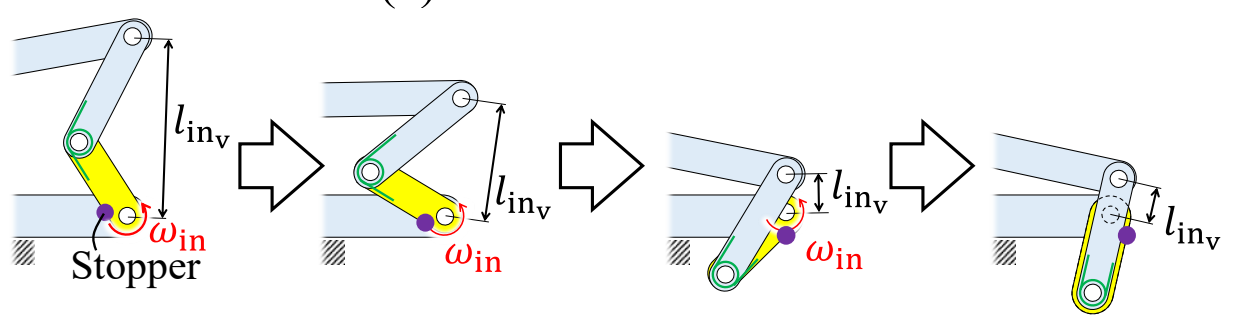

(b) Variation in variable link length mechanism when an external load is applied to the output link

**Fig. 4.** LS-CVT model

load applied to the output link causes a relative rotation of the separated input links around joint $P_3$, and $l_{\mathrm{in_v}}$ decreases. Therefore, the LS-CVT increases the output torque when an external load is applied to the output link (Fig. 4(b) shows the corresponding variation in $l_{\mathrm{in_v}}$). A stopper was installed in the input link to limit the variation in $l_{\mathrm{in_v}}$ and to stop the undesired decrease in the torque amplification ratio. Similar to the torques from the torsion spring, the forces from the stopper applied to the separated input links 1 and 2 cancel each other internally when $l_{\mathrm{in_v}}$ is minimized, and the motions of input links 1 and 2 stop. The integration of the LS-CVT and quick-return mechanisms provides the following benefits: The $l_{\mathrm{out}} \cos\theta_2 < l_{\mathrm{in_v}} \cos\theta_1$ setting amplifies the opening/closing speed of the fingers. Before the relative rotation of the input links by the stopper, the torsion spring provides high back-drivability, which can reduce the impulsive force applied to the fingers. The small motion range of the output link of the LS-CVT is significantly expanded by employing the high-speed section of the quick-return mechanism.

## III. Analyses of the Motion Speed and Tip Force

This section presents an analysis of the motion speed of the fingertip and the maximum tip force. Based on the analysis, the magnitude of the force and speed that the gripper can generate as well as the design parameters (e.g., link lengths) are determined.

### *A. Opening/closing speed of fingertips*

The closing process terminates when the fingertip comes into contact with the object. Accordingly, the opening/closing speeds of the fingertips were analyzed when no load was applied. Therefore, the trajectory of the fingertip was derived without considering the motion of the LS-CVT (i.e., only the motion of the finger unit was considered). Subsequently, the LS-CVT motion was included in the analysis, and the opening/closing speed was derived.

The nomenclature for the analysis in the first step is shown in Fig. 5. The fingertip position ($\boldsymbol{p}_{\mathrm{ft}}$) is derived for a known rotational angle ($\theta_{\mathrm{ip}}$) of the rotational plate. The output angle of the LS-CVT, $\theta_{\mathrm{ip}}$, is considered in the next step. Let $\boldsymbol{p}_{\mathrm{op}}$ and $\boldsymbol{p}_{\mathrm{ip}}$ be the center positions of the outer and inner pins, respectively; $\boldsymbol{p}_{\mathrm{c}}$ be the center position of the rotational plate; $l_{\mathrm{ft}}$ be the length of $\overline{\mathrm{P_{ft}P_{ip}}}$, $r_{\mathrm{op}}$ be the distance between $\boldsymbol{p}_c$ and $\boldsymbol{p}_{\mathrm{op}}$; and $r_{\mathrm{ip}}$ be the distance between $\boldsymbol{p}_{\mathrm{c}}$ and $\boldsymbol{p}_{\mathrm{ip}}$. The objective is to derive the relationship between position $\boldsymbol{p}_{\mathrm{ft}}$ and rotational angle $\theta_{\mathrm{ip}}$. The coordinate frame $\Sigma_A$, is set as shown in Fig. 5. Let $\phi_1$ be the angle between $\overline{\mathrm{P_{op}P_{ip}}}$ and the $x_A$ axis, and $\phi_2$ be the angle between $\overline{\mathrm{P_{op}P_{ip}}}$ and $\overline{\mathrm{P_{ft}P_{ip}}}$. Positions $\boldsymbol{p}_{\mathrm{ip}}$ and $\boldsymbol{p}_{\mathrm{op}}$ are given by

$$\boldsymbol{p}_{\mathrm{ip}} = r_{\mathrm{ip}}[\cos\theta_{\mathrm{ip}} \quad \sin\theta_{\mathrm{ip}}]^{\mathrm{T}} \tag{3}$$

$$\boldsymbol{p}_{\mathrm{op}} = [0 \quad r_{\mathrm{op}}]^{\mathrm{T}} \tag{4}$$

where $r_{\mathrm{ip}}, r_{\mathrm{op}}$, and $\phi_2$ are known constant design parameters. Let $\boldsymbol{p}_{\mathrm{op-ip}}$ be the position vector from $\boldsymbol{p}_{\mathrm{op}}$ to $\boldsymbol{p}_{\mathrm{ip}}$; $\boldsymbol{p}_{\mathrm{op-ip}}$ and $\phi_1$ are expressed as follows:

$$\boldsymbol{p}_{\mathrm{op-ip}} = [p_{\mathrm{op-ip}_x} \quad p_{\mathrm{op-ip}_y}]^{\mathrm{T}} = \boldsymbol{p}_{\mathrm{ip}} - \boldsymbol{p}_{\mathrm{op}} \tag{5}$$

$$\phi_1 = \tan^{-1}\left(p_{\mathrm{op-ip}_y}/p_{\mathrm{op-ip}_x}\right) \tag{6}$$

The fingertip position ($\boldsymbol{p}_{\mathrm{ft}}$) is obtained by the following:

$$\boldsymbol{p}_{\mathrm{ft}} = \boldsymbol{p}_{\mathrm{ip}} - l_{\mathrm{ft}}[\cos(\phi_1 - \phi_2) \quad \sin(\phi_1 - \phi_2)]^{\mathrm{T}} \tag{7}$$

From (3) to (7), $r_{\mathrm{ip}}$, $r_{\mathrm{op}}$, $l_{\mathrm{ft}}$, and $\phi_2$ are the design parameters of the finger unit. To determine these parameters, the following requirements were considered: 1) for a compact and lightweight design, the radius of the gripper was 42.5 mm; 2) the distance $\overline{\mathrm{P_{op}P_{ip}}}$ exceeded 8 mm to allow sufficient space for installing components, such as bearings; 3) the fingertip trajectory must pass through the center of the gripper, $\mathrm{P_c}$; and 4) the width of the graspable objects exceeds 70 mm (i.e., design requirement 5 described in Section II.A). Moreover, $r_{\mathrm{op}}$ was set to 35 mm to satisfy the first requirement, allowing the components to be freely dimensioned within a certain range. $l_{\mathrm{ft}}$ was set as 26.2 mm to satisfy the third requirement. The remaining parameters, i.e., $r_{\mathrm{ip}}$ and $\phi_2$, were set to maximize the speed increase, $\left\|d\boldsymbol{p}_{\mathrm{ft}}/d\theta_{\mathrm{ip}}\right\|$, which is the ratio of the quick-return mechanism. By numerically differentiating (7), the parameters were set as $r_{\mathrm{ip}}$ =26 mm and $\phi_2 = 58°$. The trajectory of $\boldsymbol{p}_{\mathrm{ft}}$ and speed increase, $\left\|d\boldsymbol{p}_{\mathrm{ft}}/d\theta_{\mathrm{ip}}\right\|$, derived from the set parameters are shown in Fig. 6. The $\boldsymbol{p}_{ft}$ can reach up to 42 mm from the $\mathrm{P_c}$, and $\theta_{\mathrm{ip}}$ ranges from 74.5° to 105°. The angle when the gripper is fully closed (i.e., $\theta_{\mathrm{ip}}$=74.5°) is the angle when $\|\boldsymbol{p}_{\mathrm{ft}}\| =$ 3.5 mm; this is because the fingertip is cylindrical in shape with a 3.5 mm radius (see Fig. 6(a)). Thus, the graspable range of the gripper was calculated as 77 mm (=$(42 - 3.5) \times 2$). As shown in Fig. 6(b), the speed increase ratio of the quick-return mechanism was maximized in this $\theta_{\mathrm{ip}}$ range.

Next, the speed-increasing ratio of the LS-CVT was derived when no external load was applied to the output link. The model and nomenclature used for the analysis are shown in Fig. 7. Here, the rotational angle of the output link of the LS-CVT, $\theta_{\mathrm{out}}$, is derived when the angle of the input link attached to the motor axis $\theta_{\mathrm{in_v}}^{no-contact}$ (angle between the virtual input link

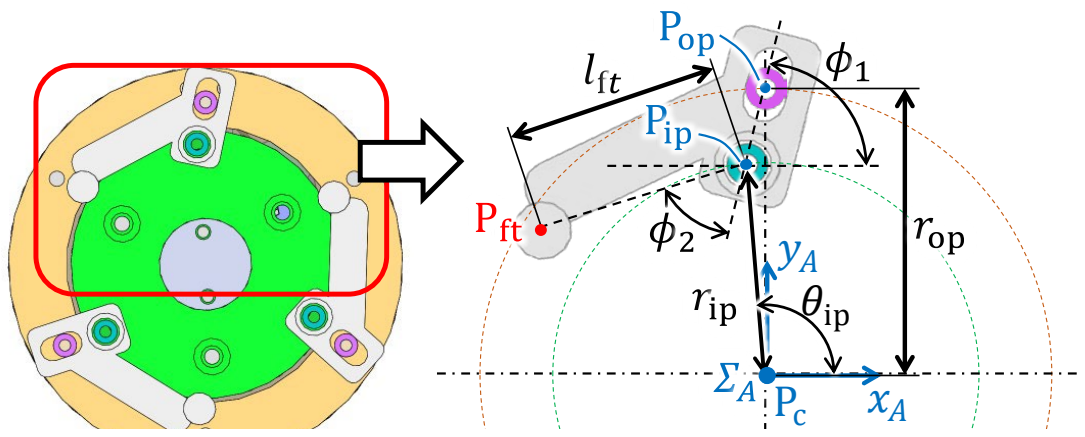

**Fig. 5.** Model of the quick-return mechanism for analyzing the motion speed

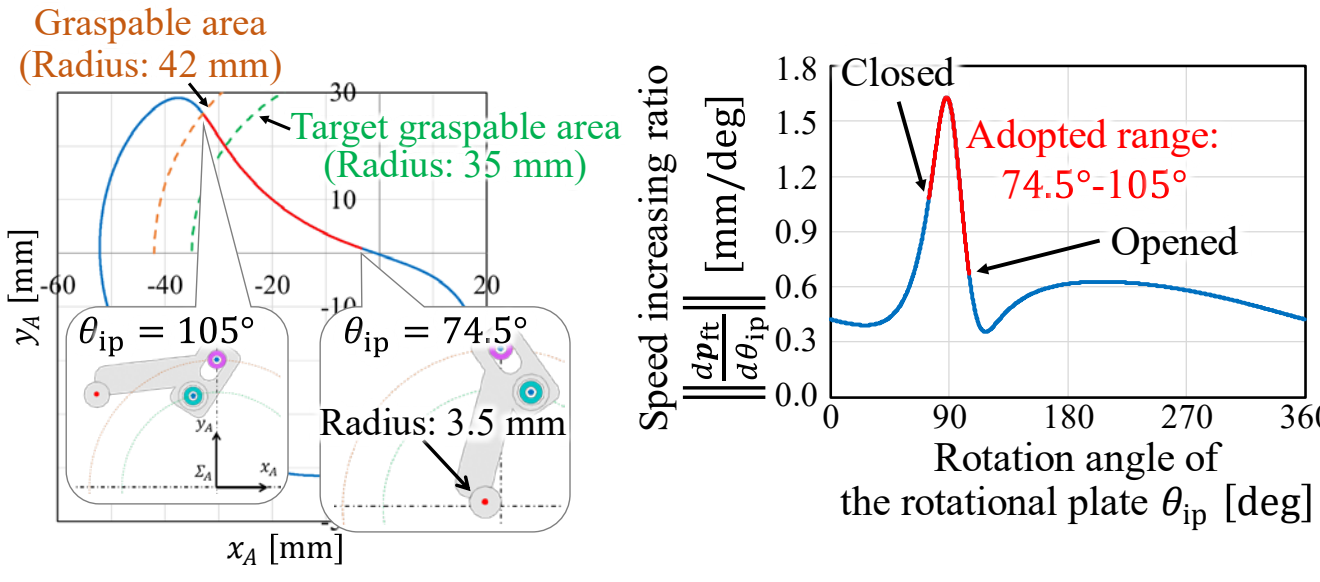

(a) Trajectory (b) Speed-increasing ratio

**Fig. 6.** Analysis results of the quick-return mechanism

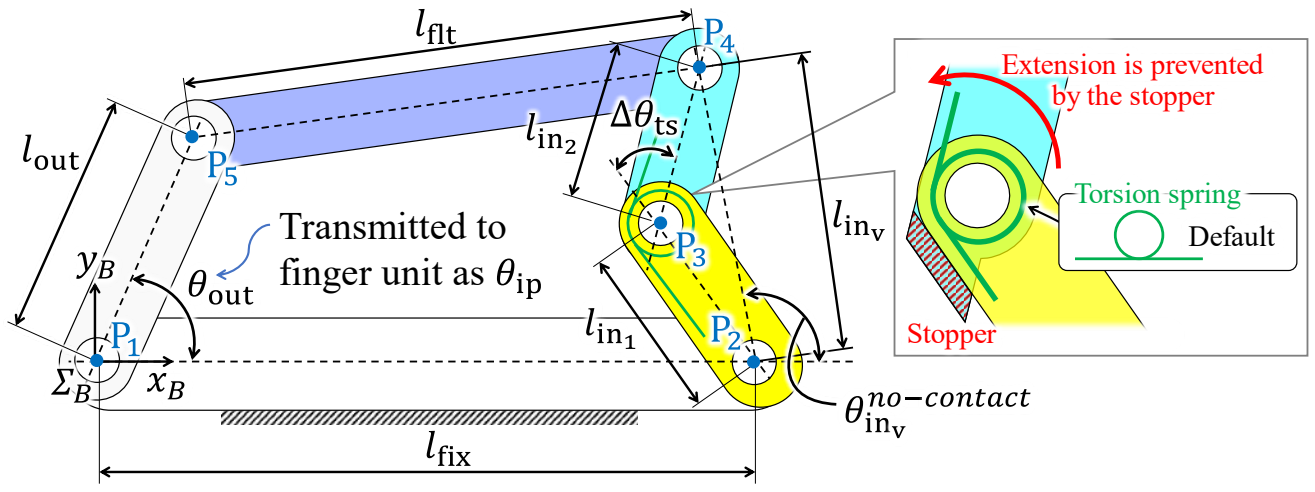

**Fig. 7.** Model of the LS-CVT for analyzing the motion speed of the fingertip

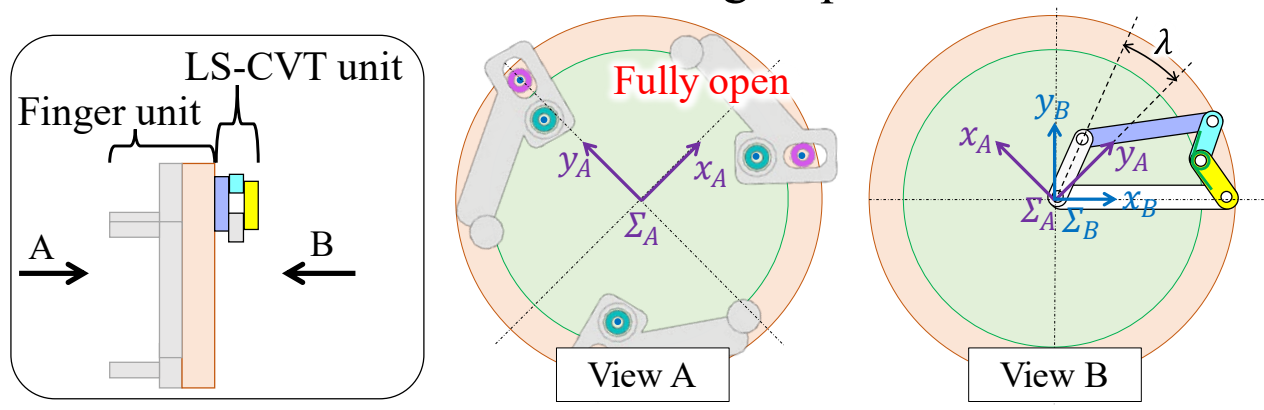

**Fig. 8.** Positional relationship between the finger and LS-CVT units

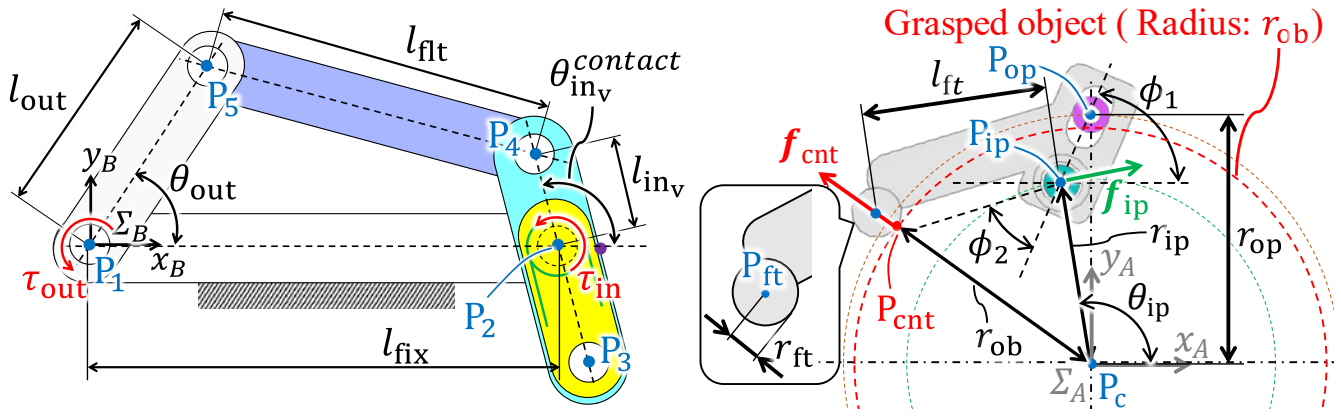

(a) LS-CVT (b) Quick-return mechanism

**Fig. 9.** Model for analyzing the maximum tip force

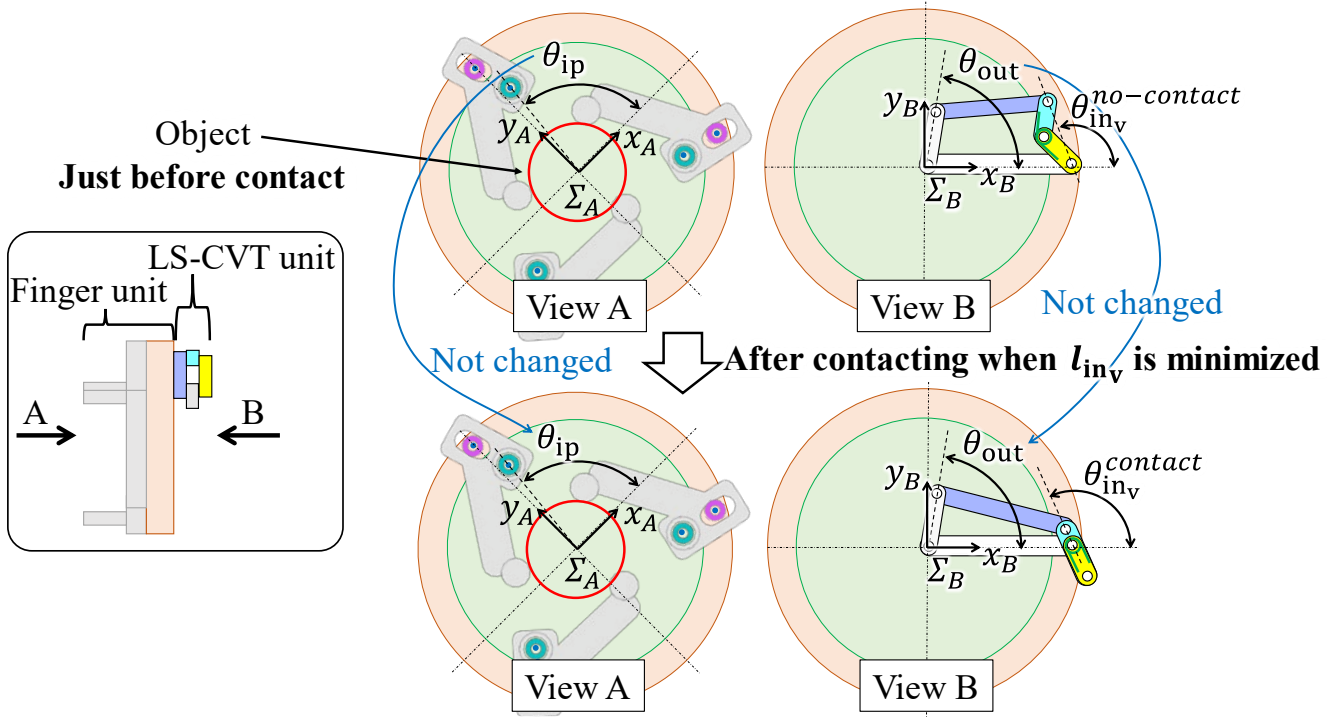

**Fig. 10.** Schematic of the state immediately before the fingertip contacts the object and the state in which the length of the virtual input link is minimized after contact

and $x_B$ axis), is given. The superscript '*no-contact*' refers to the angle at which the fingertip does not make contact with the object. Let $\Delta\theta_{ts}$ be the angle between links $\overline{P_2P_3}$ and $\overline{P_3P_4}$ at the initial state where no external load is applied to the LS-CVT. As described in Section II.B.2, the stopper exists in the input links to maintain the minimized $l_{in_v}$ after the fingertips make contact with the object. In addition, another stopper was also installed between the input links to maintain $\Delta\theta_{ts}$ during the opening and closing operations. The stopper prevents the relative rotation between the input links in the direction in which the angle between links $\overline{P_2P_3}$ and $\overline{P_3P_4}$ becomes smaller than $\Delta\theta_{ts}$. As the default leg angle of torsion spring installed between the input links is 180°, the torsion spring was set to be preloaded by the displacement of $\Delta\theta_{ts}$ in the initial state of the LS-CVT. Using the stopper and preloading, the angle between links $\overline{P_2P_3}$ and $\overline{P_3P_4}$ can be maintained during the opening and closing operations. In this study, $\Delta\theta_{ts}$ was set to 45°, which was the minimum angle that can prevent the relative rotation between the input links during the opening and closing operations. The minimization is also equivalent to maximizing the amplification ratio of the angular velocity because decreasing $\Delta\theta_{ts}$ is equivalent to increasing $l_{in_v}$. The coordinate frame $\Sigma_B$, is set as shown in Fig. 7. The origin of $\Sigma_B$ is located at $P_1$ (the output joint). As described in Section II.B.2, the virtual input link was considered for the analysis. As no load was applied to the fingertip, $l_{in_v}$ remained constant. From the geometrical relationship, the relationship between $\theta_{out}$ and $\theta_{in_v}^{no-contact}$ is

$$\theta_{out} = \cos^{-1}\frac{\alpha_2\alpha_3 - \alpha_1\sqrt{\alpha_1^2+\alpha_2^2-\alpha_3^2}}{\alpha_1^2+\alpha_2^2} \tag{8}$$

where

$$\alpha_1 = 2l_{in_v}l_{out}\sin\theta_{in_v} \tag{9}$$

$$\alpha_2 = 2l_{in_v}\left(l_{fix} + l_{out}\cos\theta_{in_v}^{no-contact}\right) \tag{10}$$

$$\alpha_3 = l_{out}^2 - l_{flt}^2 + l_{in_v}^2 + l_{fix}^2 - 2l_{out}l_{fix}\cos\theta_{in_v}^{no-contact} \tag{11}$$

By numerically differentiating (8), the speed-increasing ratio $d\theta_{out}/d\theta_{in_v}^{no-contact}$, is derived. The speed-increasing ratio nearly doubled within the adopted range. The following parameter values were used: $l_{in_1}$= 8.5 mm, $l_{in_2}$= 11 mm, $l_{fix}$= 23 mm, $l_{flt}$ = 24 mm, $l_{out}$ = 9 mm, and $\lambda$ = 4.3°, where $\lambda$ represents the difference between $\theta_{out}$ (output angle of LS-CVT) and $\theta_{ip}$ (input angle of the rotational plate of the finger unit) (Fig. 8):

$$\theta_{out} = \theta_{ip} + \lambda. \tag{12}$$

The next section explains how the link lengths and $\lambda$ are determined to analyze the tip force. From (7) and (8), the fingertip velocity was derived assuming that the angular velocity of the motor, $\dot{\theta}_{in_v}^{no-contact}$, was 412 °/s (=68.7 rpm: measured maximum angular velocity of the motor under no-load conditions). The calculated maximum speed of motion was 1410 mm/s.

### *B. Tip force*

The tip force reaches its maximum when the length of the virtual input link, $l_{in_v}$ (see Fig. 4(b)) is minimized. This minimum length occurs when an external load is applied to the fingertip (i.e., the output link of the LS-CVT). Accordingly, the tip force is considered in this situation. The model and nomenclature used for the analysis are shown in Fig. 9. Let $\theta_{in_v}^{contact}$ be $\theta_{in_v}$ when the fingertip is in contact with an object, and $l_{in_v}$ is minimized. Moreover, Fig. 10 illustrates the states immediately before contact and when the length of the virtual input link is minimized owing to contact. $\theta_{out}$ and $\theta_{ip}$ do not change owing to the contact because the fingertip positions are maintained. Their values are determined based on the radius of the grasped object.

First, we consider the relationship between the motor input torque $\tau_{in}$, and the LS-CVT output torque $\tau_{out}$. Letting $l_{in_1}$ and $l_{in_2}$ be the lengths of the two separated input links, the minimized $l_{in_v}$ corresponds to $|l_{in_2} - l_{in_1}|$(see Fig. 9(a)). Let $\boldsymbol{e}_{n45}$ be the unit vector representing the normal direction of segment $\overline{\mathrm{P_4P_5}}$ (the direction toward the origin is positive). The output torque ($\tau_{\mathrm{out}}$) is expressed as

$$\tau_{\mathrm{out}} = \varepsilon_{\mathrm{amp}}\tau_{\mathrm{in}} = \left|\frac{\boldsymbol{e}_{n_{45}}^{\mathrm{T}}\boldsymbol{p}_{15}}{\boldsymbol{e}_{n_{45}}^{\mathrm{T}}\boldsymbol{p}_{24}}\right|\tau_{\mathrm{in}} \tag{13}$$

where

$$\boldsymbol{e}_{n_{45}} = \frac{1}{l_{flt}}[-\boldsymbol{p}_{45_y} \quad \boldsymbol{p}_{45_x}]^{\mathrm{T}} \tag{14}$$

$$\boldsymbol{p}_{15} = l_{\mathrm{out}}[\cos\theta_{\mathrm{out}} \quad \sin\theta_{\mathrm{out}}]^{\mathrm{T}} \tag{15}$$

$$\boldsymbol{p}_{24} = l_{\mathrm{in_v}}\left[\cos\theta_{\mathrm{in_v}}^{contact} \quad \sin\theta_{\mathrm{in_v}}^{contact}\right]^{\mathrm{T}} \tag{16}$$

$\varepsilon_{\mathrm{amp}}$ is the torque amplification ratio.

The design parameters, including the link lengths ($l_{\mathrm{in_1}}$, $l_{\mathrm{in_2}}$, $l_{\mathrm{fix}}$, $l_{\mathrm{flt}}$, and $l_{\mathrm{out}}$) and the angle between the finger unit and LS-CVT, $\lambda$, were derived under the following conditions: 1) the displacement range of $\theta_{\mathrm{out}}$ is 30.5° (=105°–74.5°), as explained in Section III.A; 2) the LS-CVT must be placed inside a circular area with a radius of 35 mm when viewed from the center of the gripper to avoid interference among the component parts; and 3) the displacement ranges of $\theta_{\mathrm{in_v}}^{non-contact}$ and $\theta_{\mathrm{in_v}}^{contact}$ do not include the dead point. While satisfying the aforementioned conditions, the design parameters were determined through a grid search to achieve the following objectives: 1) the ratio of the output link length to the minimized virtual input link length must be as large as possible to maximize the torque amplification ratio, and 2) the ratio of the output link length to the maximized virtual input link length must be as small as possible to maximize the speed-increasing ratio.

As described in Section III.A, the link lengths obtained were $l_{\mathrm{in_1}}$ = 8.5 mm, $l_{\mathrm{in_2}}$= 11 mm, $l_{\mathrm{fix}}$ = 23 mm, $l_{\mathrm{flt}}$ = 24 mm, and $l_{\mathrm{out}}$= 9 mm. From the link lengths, $\lambda$ = 4.3° was obtained analytically. Moreover, Fig. 11 shows the representative results, including those obtained using the adopted parameters, of the amplification ratio ($\varepsilon_{\mathrm{amp}}$) calculated using (13) when varying the design parameters that satisfy the three conditions above. The adopted parameters provide a large amplification ratio throughout the targeted range of $\theta_{\mathrm{in_v}}^{contact}$. The LS-CVT can amplify the input torque from the motor more than three times in the operating range.

Next, we consider the relationship between the tip force ($\boldsymbol{f}_{\mathrm{cnt}}$) and input torque to the rotational plate equal to the output torque of the LS-CVT ($\tau_{\mathrm{out}}$). Fig. 9(b) shows an analytical model for grasping a cylindrical object with radius $r_{\mathrm{ob}}$. Let $\mathrm{P_{cnt}}$ be the contact point between the object and fingertip, $\boldsymbol{f}_{\mathrm{cnt}}$ be the contact force corresponding to the tip force, and $\boldsymbol{f}_{\mathrm{ip}}$ be the force applied to the inner pin by the rotation of the rotational plate.

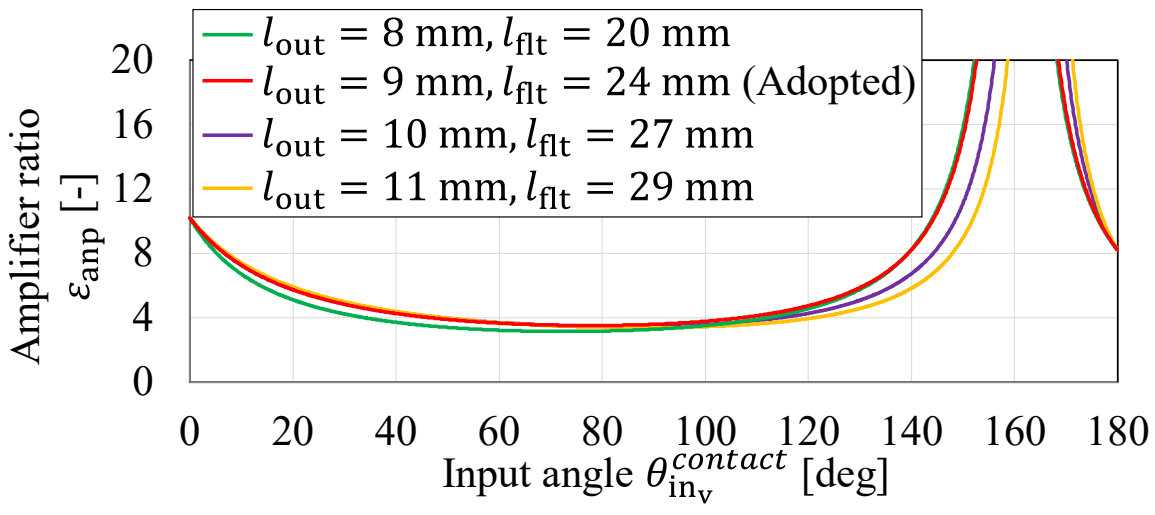

**Fig. 11.** Amplification ratio of output to input torques with varying link lengths ( $l_{\mathrm{out}}$ and $l_{\mathrm{flt}}$).

The equilibrium of moments around $\mathrm{P_{op}}$ is given by

$$\boldsymbol{p}_{\mathrm{op-cnt}} \times \boldsymbol{f}_{\mathrm{cnt}} + \boldsymbol{p}_{\mathrm{op-ip}} \times \boldsymbol{f}_{\mathrm{ip}} = \boldsymbol{0} \tag{17}$$

where $\boldsymbol{p}_{\mathrm{op-cnt}}$ and $\boldsymbol{p}_{\mathrm{op-ip}}$ denote the position vectors representing points $\mathrm{P_{cnt}}$ and $\mathrm{P_{ip}}$ from $\mathrm{P_{op}}$, respectively, and $\boldsymbol{p}_{\mathrm{op-ip}}$ is given by (5). Assuming that the contact force is applied only along the radial direction of the object to derive the maximum tip force, $\boldsymbol{f}_{\mathrm{cnt}}$ and $\boldsymbol{p}_{\mathrm{op-cnt}}$ are expressed as

$$\boldsymbol{f}_{\mathrm{cnt}} = \frac{\|\boldsymbol{f}_{\mathrm{cnt}}\|}{\|\boldsymbol{p}_{\mathrm{ft}}\|}\boldsymbol{p}_{\mathrm{ft}}; \tag{18}$$

$$\boldsymbol{p}_{\mathrm{op-cnt}} = \boldsymbol{p}_{\mathrm{cnt}} - \boldsymbol{p}_{\mathrm{op}} = \frac{r_{\mathrm{ob}}}{\|\boldsymbol{p}_{\mathrm{ft}}\|}\boldsymbol{p}_{\mathrm{ft}} - \boldsymbol{p}_{\mathrm{op}}. \tag{19}$$

From the geometrical relationship, the radius, $r_{\mathrm{ob}}$, of the object is expressed by

$$r_{\mathrm{ob}} = \|\boldsymbol{p}_{\mathrm{ft}}\| - r_{\mathrm{ft}} \tag{20}$$

where $r_{\mathrm{ft}}$ denotes the radius of the cylindrical fingertips. The acting direction of $\boldsymbol{f}_{\mathrm{ip}}$ is vertical to the segment $\overline{\mathrm{P_cP_{ip}}}$ and toward the counterclockwise direction of the rotating disk. Thus, $\boldsymbol{f}_{\mathrm{ip}}$ can be obtained as follows:

$$\boldsymbol{f}_{\mathrm{ip}} = \frac{\|\boldsymbol{f}_{\mathrm{ip}}\|}{\|\boldsymbol{p}_{\mathrm{ip}}\|}\begin{bmatrix} p_{\mathrm{ip}_y} \\ -p_{\mathrm{ip}_x}\end{bmatrix} = \|\boldsymbol{f}_{\mathrm{ip}}\|\begin{bmatrix}\sin\theta_{\mathrm{ip}} \\ -\cos\theta_{\mathrm{ip}}\end{bmatrix} \tag{21}$$

The magnitude of $\boldsymbol{f}_{\mathrm{ip}}$ is given as follows:

$$\|\boldsymbol{f}_{\mathrm{ip}}\| = \tau_{\mathrm{out}}/r_{\mathrm{ip}} \tag{22}$$

From (17) and (20), the contact magnitude of the maximum tip force $\|\boldsymbol{f}_{\mathrm{cnt}}\|$, is derived if $r_{\mathrm{ob}}$ is given. Assuming that the torque $\tau_{\mathrm{in}}$, of the motor is 1300 N·mm, the maximum tip force that can be generated in the developed gripper can theoretically exceed 50 N. The details of the theoretical tip force, including its profile, are provided along with the experimental results in Section III.C.2.

### *C. Experimental validation*

This section presents an experimental investigation of the motion speed and maximum tip force of the developed gripper to validate the above analyses. Moreover, the impulsive force when making contact between the finger and the object was also evaluated because the impulsive force could be large owing to the high-speed grasping motion. In the experiments described in this section, the motor was operated with current control, and the desired motor current was set to its maximum. In this setting, the motor velocity corresponded to the maximum velocity because no external load was applied to the fingertip during the opening and closing motions.

#### *1) Motion speed of the fingertip*

The speed and trajectory of the fingertip were measured when the finger was closed. The experimental setup and results are shown in Fig. 12. As shown in Fig. 12(a), the bottom surface of the fingertip was colored red, and its motion was monitored using a high-speed camera placed below the gripper. The opening and closing operations were conducted three times, and the trajectory of the fingertip was obtained by processing the captured camera image. The motion speed was calculated from the measured fingertip trajectory. Fig. 12(b) shows the measured and theoretically derived trajectories (Section III.A). The measured trajectory was virtually identical to the theoretical value. The mean maximum motion speed was 1396 mm/s, whereas the theoretical maximum motion speed was 1410 mm/s (Section III.A). The measured and theoretical values of the maximum speed are extremely close to each other. In addition, the time from the fully open state to the fully closed state was measured by extracting snapshot frames from the moment the motion began until it stopped. The mean value of the measured closing time was 62 ms. In summary, the analysis results of the motion speed were reasonable, and the developed gripper generated a high motion speed.

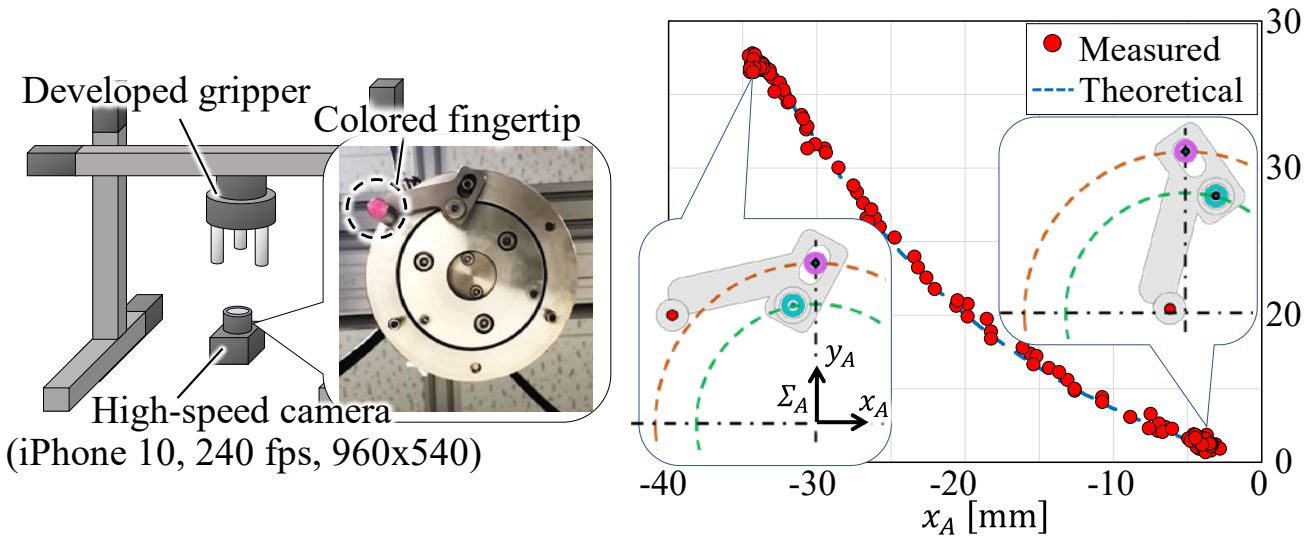

(a) Experimental setup (b) Measured trajectory

**Fig. 12.** Measurement of the finger speed and trajectory

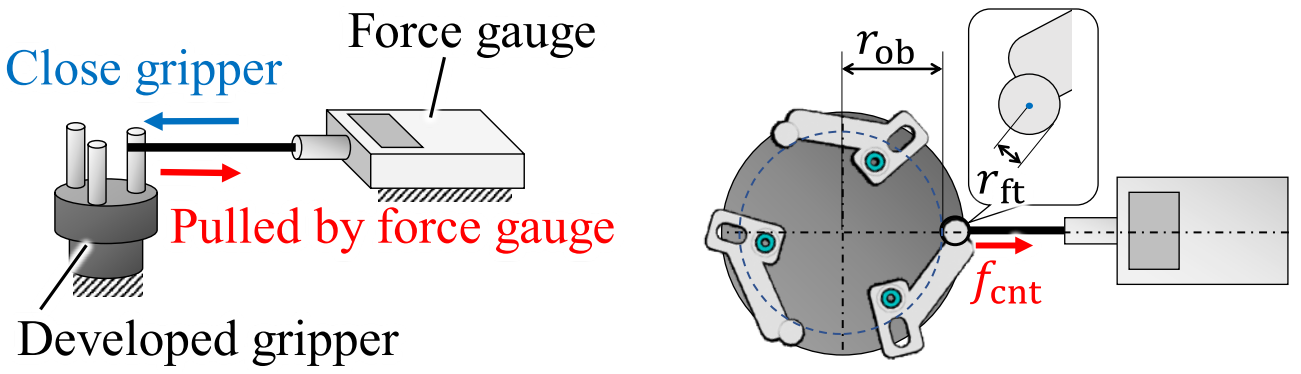

(a) Whole view (b) Top view

**Fig. 13.** Experimental setup for measuring the tip force

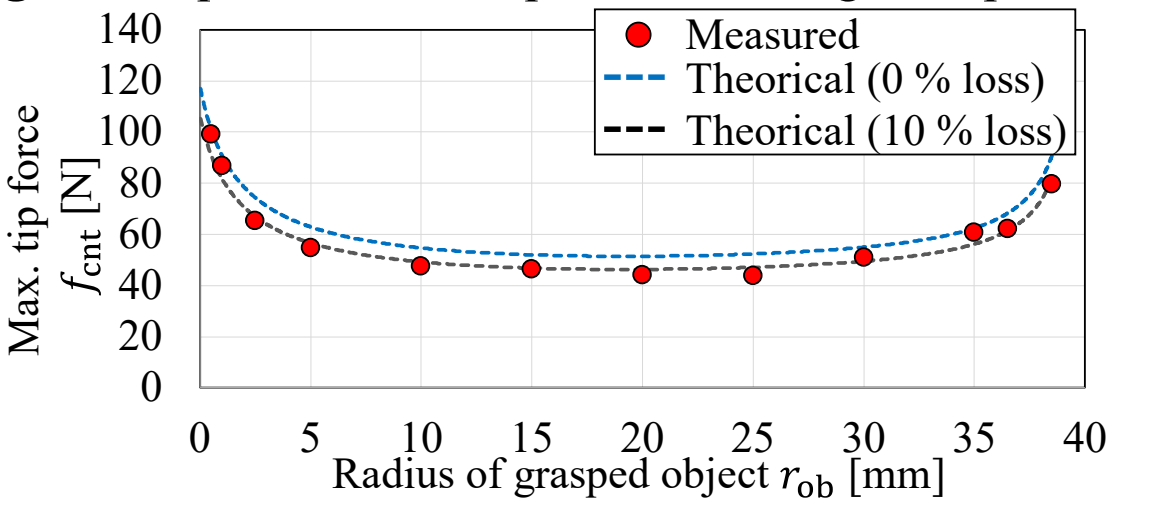

**Fig. 14.** Measured and theoretical maximum tip forces

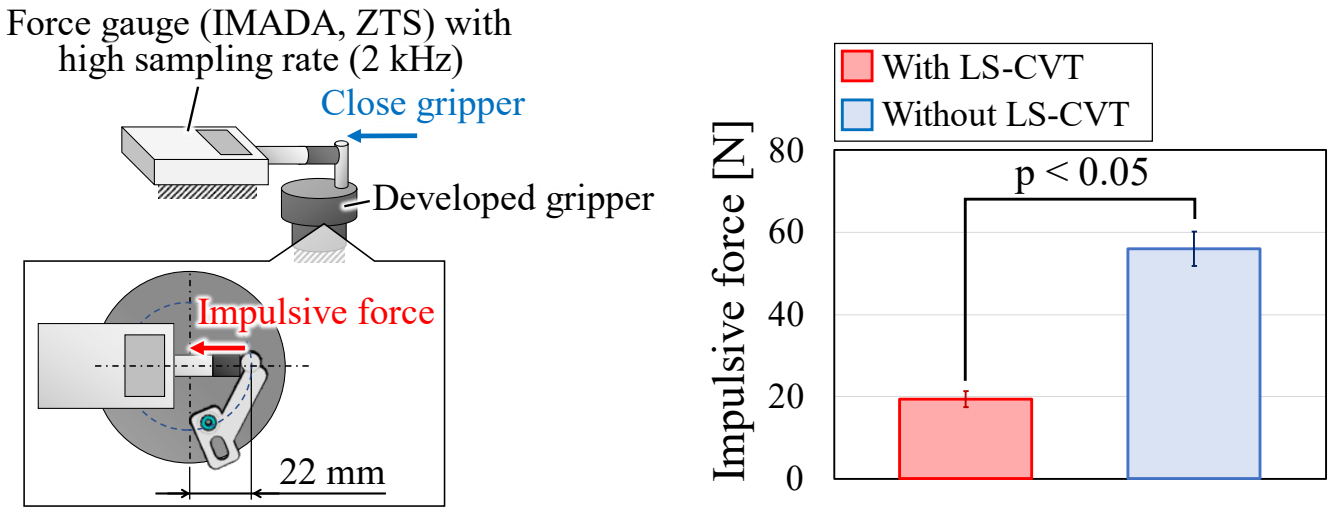

(a) Experimental setup (b) Result

**Fig. 15.** Experimental setup and result of the impulsive force

*2) Maximum tip force*

This section presents an evaluation of the maximum tip force. Fig. 13 shows the experimental setup. The fingertip of the gripper was connected to a force gauge (IMADA, ZTA-500N) that measured the pulling force when the gripper was closed. In the experiment, the gripper and force gauge were placed such that the action line of the pulling force crossed the center of the gripper, and the distance from the center of the gripper to the fingertip was $r_{ob} + r_{ft}$. With this setting, the pulling force, measured by the force gauge, corresponds to the magnitude of the maximum tip force, $f_{cnt}$, when an object with a width of $2r_{ob}$ is grasped by the gripper. The tests were conducted three times for each $r_{ob}$ (from 0.5 to 38.5 mm (maximum graspable radius)). The measured and theoretical forces (described in Section III. B) are shown in Fig. 14. The experimental values are slightly lower than the theoretical values. Assuming that a 10 % loss exists because of friction (i.e., theoretical value $\times 0.9$), they are virtually the same. This confirms that the analysis of the maximum tip force is valid, and that the developed gripper can generate a high tip force.

*3) Impulsive force at contact*

The impulsive force when the finger made contact with an object was evaluated. The experimental setup is shown in Fig. 15(a). From (7) and (8), the fingertip speed was at its maximum when the finger started moving from the fully opened state and moved 22 mm away from the center of the gripper. The impulsive force was measured as the finger passed through that position. The experiments were conducted ten times for each motor velocity. For comparison, the same experiments were conducted under the condition where input links 1 and 2 of the LS-CVT were connected such that they did not rotate relatively; that is, the LS-CVT was replaced by a conventional four-link mechanism (we call this condition "without LS-CVT"). The results are shown in Fig. 15(b). The magnitude of the impulsive force with the LS-CVT is significantly smaller than that without the LS-CVT. The LS-CVT could absorb the impulsive force owing to the dumping effect of the torsion spring until $l_{in_v}$ reached a minimum. Notably, owing to the absorbing function, the impulsive force was smaller than the maximum tip force, as shown in Fig. 14.

## IV. Fundamental Grasp Test and Grasping Strategy

This section presents grasping tests for primitive-shaped objects and the formulation of a grasping strategy.

### A. Grasping test for primitive-shaped objects

To evaluate the grasping ability of the developed gripper, grasping tests were conducted on primitive-shaped objects. The target shapes are cylindrical, rectangular, and triangular. The dimensions of the target objects are listed in Table I. Nine types of 3D-printed objects with different bottom-face shapes were prepared. Fig. 16 illustrates the experimental setup. A high-speed camera was used to monitor the grasping state. Initially, the target object was randomly placed at a position that was within the graspable range of the gripper (i.e., within 38.5 mm from the center of the gripper). A grasping operation was then implemented to determine whether the grasping was successful. The cylindrical objects were grasped using the self-centering function with a displacement error between the initial position of the target object and the center of the gripper, as investigated in [6].

Table I Sizes of Target Objects

| | | Cylinder | | | Rectangular | | | Triangular | | |
|---|---|---|---|---|---|---|---|---|---|---|
| Object No. | | 1 | 2 | 3 | 4 | 5 | 6 | 7 | 8 | 9 |
| Dimensions [mm] | $l_1$ | 10 | 50 | 70 | 30 | 70 | 30 | 30 | 30 | 30 |
| | $l_2$ | - | - | - | 30 | 70 | 50 | 30 | 50 | 50 |
| | $l_3$ | - | - | - | - | - | - | 30 | 50 | 70 |
| | Height | | | | | 10 | | | | |
| Dimensional definition | | 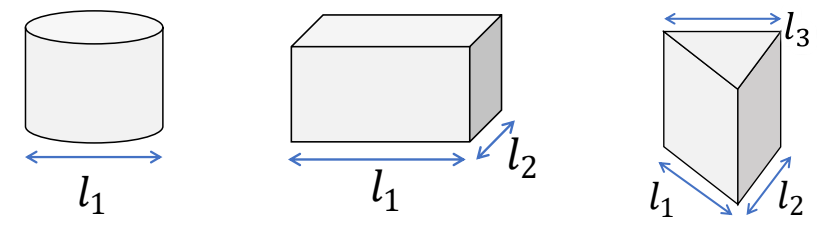 | | | | | | | | |

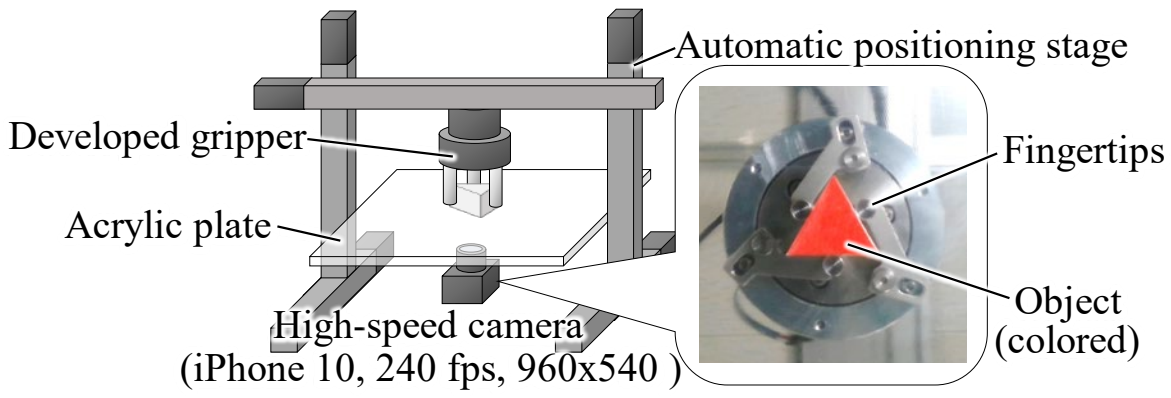

**Fig. 16.** Experimental setup for fundamental grasping tests

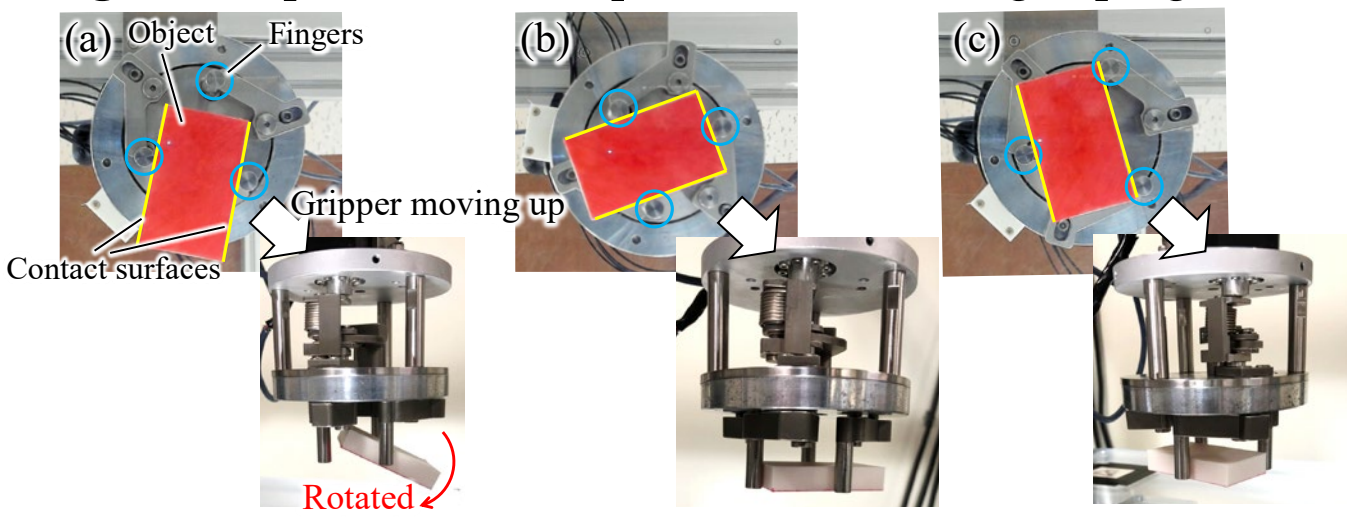

**Fig. 17.** Grasping state when two or three fingers are in contact with parallel surfaces. (a) Only two fingers are in contact with the surface. (b) Three fingers make contact with different surfaces. (c) Two fingers are in contact with the same surface and the other finger is in contact with parallel and opposite surfaces.

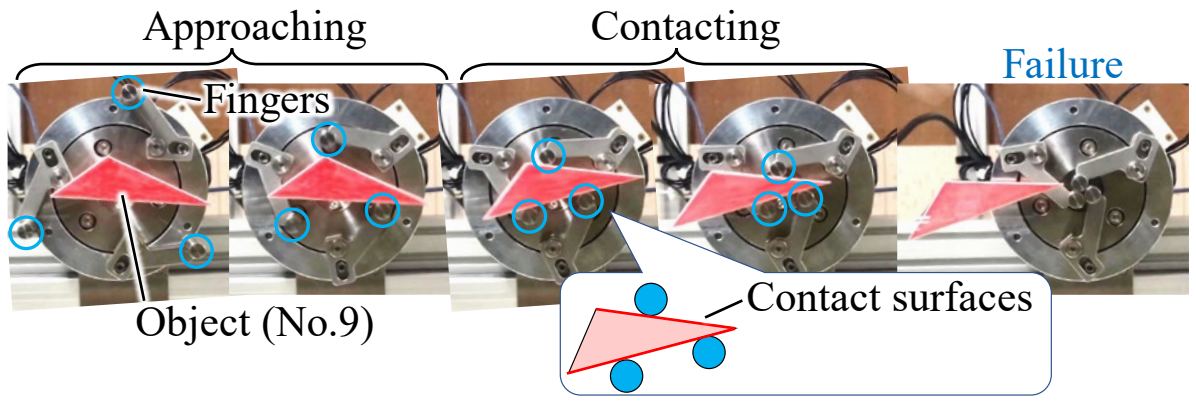

**Fig. 18.** Snapshots when grasping triangular objects (No. 9)

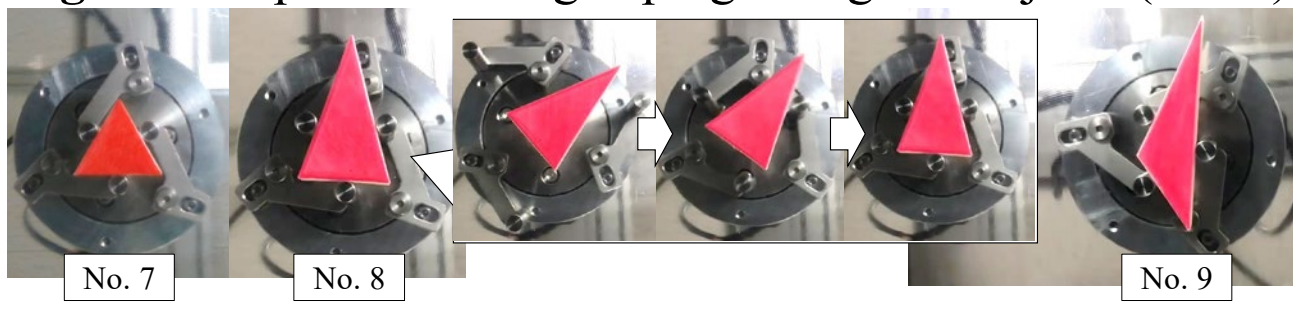

**Fig. 19.** Grasping state when grasping triangular objects and snapshot when grasping object No. 8; the self-centering function was effective

Three case types were observed when grasping rectangular objects, as shown in Fig. 17. The case in which two fingers are in contact with different parallel surfaces of the object and one finger is not is shown in Fig. 17(a). In this case, the rotation around the line connecting the two contact points cannot be constrained; hence, the grasped object can rotate around the line when lifted. A case in which all fingers are in contact with different surfaces and force-closure grasping [15] is formed is shown in Fig. 17(b). The forces at the two antipodal contact points contribute significantly to the grasping, whereas the contribution of the force at the other contact point was negligible. Therefore, the rotation shown in Fig. 17(a) can occur when the contact point generating the low-magnitude force is detached owing to a disturbance. A case in which two fingers are in contact with the same surface and the other finger is in contact with opposite and parallel surfaces is shown in Fig. 17(c). In this case, all fingers contribute significantly to grasping, and the possibility of fingertip detachment owing to external disturbances is low. Therefore, this configuration is preferable.

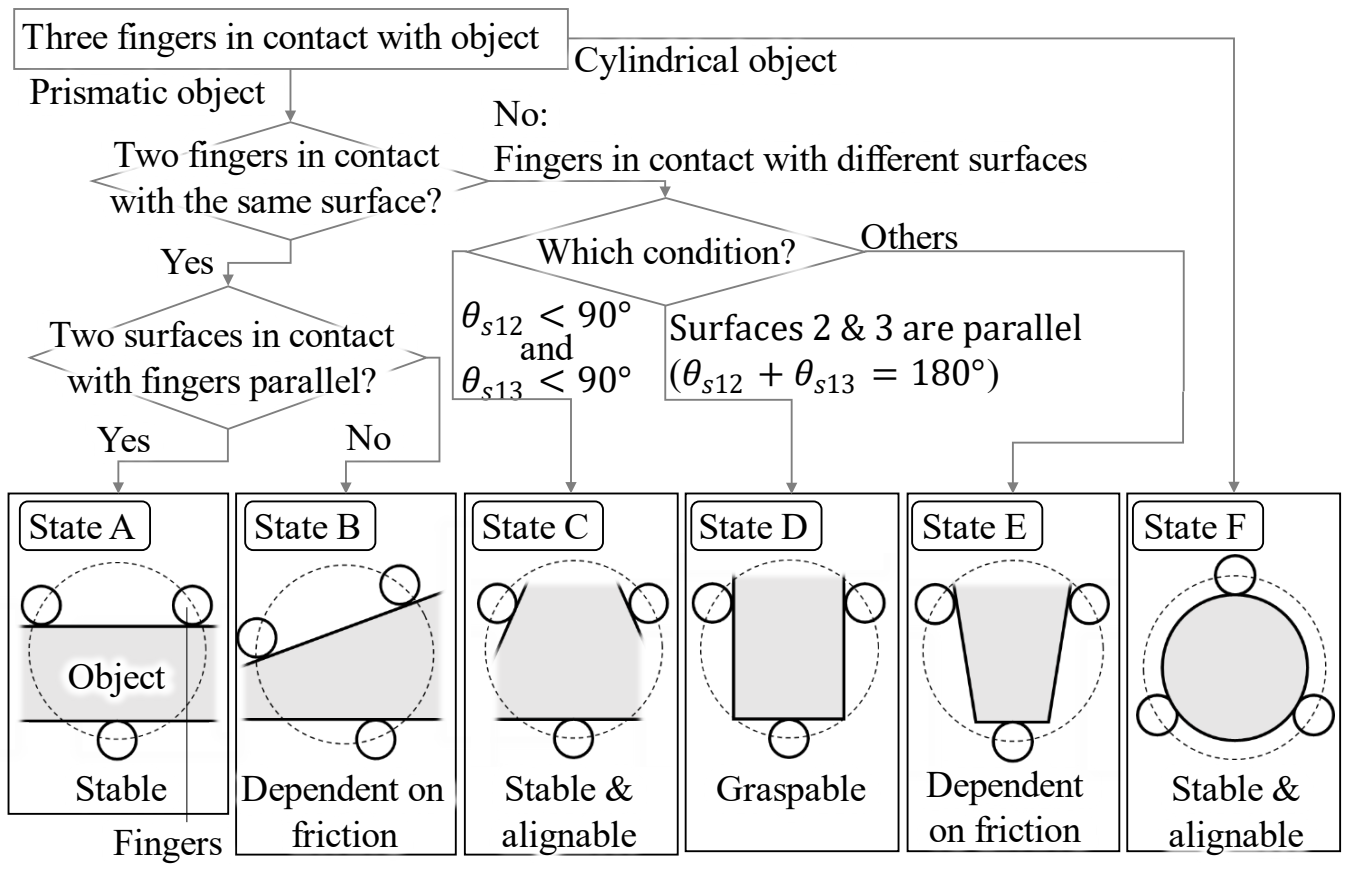

**Fig. 20.** Case analysis of the grasping state for the grasping strategy

Grasping failure was observed when the gripper attempted to grasp a triangular object. An example of this failure (target object No. 9) is shown in Fig. 18. The two fingers were in contact with the same surface, and the other finger was in contact with the opposite surface. The angle between the two surfaces was insufficient to provide antipodal forces to achieve a force-closure grasp. By contrast, if the object was located such that the three fingers were in contact with different surfaces, grasping was successful, as shown in Fig. 19. The observed configurations are only those shown in Figs. 18 and 19 for the triangular tests. When grasping triangular objects, it is preferable to have the three fingers in contact with different surfaces. The self-centering function is also valid for triangular objects if a grasping configuration similar to that shown in Fig. 19 can be achieved. In this case, the object is always grasped using the same configuration.

The grasping configurations were classified based on the observed results, as shown in Fig. 20. The grasping positions always constitutes a regular triangle. Therefore, the geometrical relationship between the regular triangle and the object surfaces determines the grasping configuration. Regarding prismatic objects, if two fingers are in contact with the same surface of the object, the grasping configuration is either of the configurations shown in Figs. 17(c) and 18. The difference between them is the angle between the two surfaces in contact with the fingers. If the two surfaces are parallel, the configuration is referred to as “State A”: otherwise, it is referred to as “State B.” In State A, the directions of the internal grasping forces are equal to the normal contact directions on both surfaces; thus, force closure was ensured. By contrast, in State B, the larger the angle between the normal directions of the contact surfaces, the more difficult it becomes to achieve force closure. The angle between the two surfaces that provides stable grasping (force closure) depends on the frictional conditions and object weight. Hence, State A is preferred, and state B is not.

When the three fingers are in contact with different surfaces, the state is one of those shown in Figs. 17(b) and 19. The grasping states are classified based on the geometrical relationships among the three surfaces. A model in which all the fingers are in contact with the surfaces is illustrated in Fig. 21. Here, without a loss of generality, grasping in a planar space, including the bottom face, is discussed. Moreover, as shown in Fig. 21(a), the following are considered: the sides where the fingers are in contact form a triangle, and the direction of the side with which finger 1 is in contact is set as the horizontal direction. If the sides do not intersect, a triangle is formed by virtually extending the side lengths, thus allowing intersections to be constructed. A triangle with a dotted line that expands the triangle with a solid line by the radius ($r_{\mathrm{ft}}$) of the cylindrical part of the fingertip is considered. Let $\theta_{sij}$ be the angle between the sides $i$ and $j$ ($i, j \in \{1,2,3\}$). If $\theta_{s12} < 90°$ and $\theta_{s13} < 90°$, the grasp configuration is State C. If sides 2 and 3 are parallel, $\theta_{s12} + \theta_{s13} = 180°$ (see Fig. 22), and the grasp configuration is State D. Otherwise, the grasp configuration is State E. The self-centering function works when the grasping configuration is in State C. If the target triangular object is of a shape that can constitute State C and a grasping strategy is applied such that the grasping configuration goes to State C (see the next subsection), the self-centering movement of the triangular object can be achieved regardless of the initial position and orientation of the triangle.

When the target object is a cylinder, the configuration is the only one in which all three fingers contact the object symmetrically. The self-centering function operates. The state refers to State F and can be achieved when the object is placed within the graspable range of the gripper [6]. Hence, we discuss the grasping states of prismatic objects because a special strategy for grasping cylindrical objects is not required.

Here, we focused on State C to observe the details of grasping stability in this state. Let $\mathrm{P}_{ci}$ be the $i$-th contact point on side $i$, and $\boldsymbol{n}_{ci}$ be the inward unit normal on contact side $i$. A set of vectors, $\boldsymbol{v}_1, \cdots, \boldsymbol{v}_k$ ($\in \mathbb{R}^2; k \geq 3$), positively spans $\mathbb{R}^2$ if any vector, $\boldsymbol{v} \in \mathbb{R}^2$, can be written as $\boldsymbol{v} = \sum_{i=1}^{k} \beta_i \boldsymbol{v}_i$ ($\beta_i \geq 0, \forall i$) [16]. If $\boldsymbol{n}_{c1}, \boldsymbol{n}_{c2}$, and $\boldsymbol{n}_{c3}$ positively span $\mathbb{R}^2$, then the object is immobilized in any translational direction. This condition is valid for State C but not for States D and E. The lemma given in [16], including the target situation, has been modified.

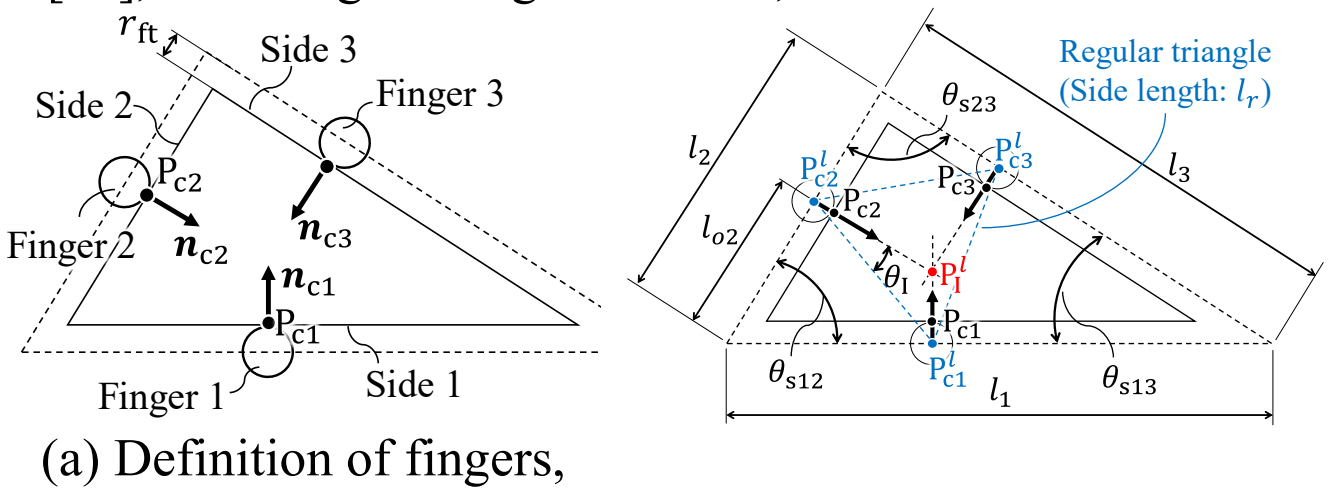

(a) Definition of fingers, surfaces, contact points, and normal forces (b) Additional information

**Fig. 21.** Model for grasping object with three fingers

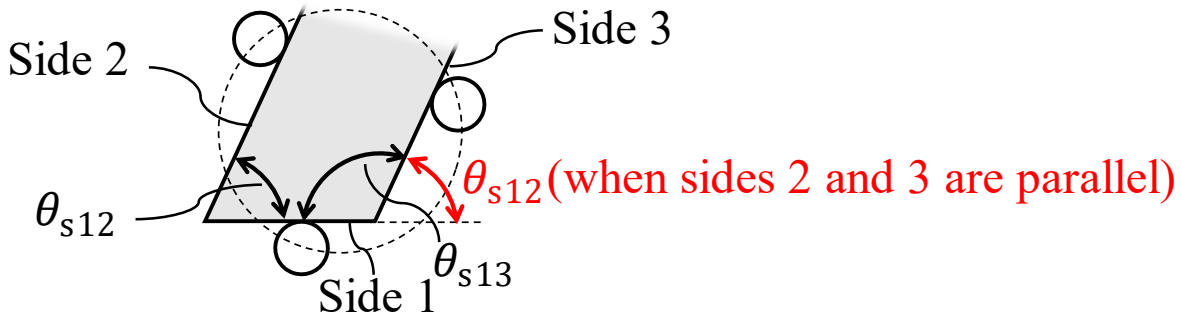

**Fig. 22.** Case in which sides 2 and 3 are parallel. This condition can be expressed as $\theta_{s12} + \theta_{s13} = 180°$.

**Lemma 1** *Contact points* $\mathrm{P}_{c1}, \mathrm{P}_{c2}$, and $\mathrm{P}_{c3}$ *achieve the second-order immobility of the object if* $\boldsymbol{n}_{c1}, \boldsymbol{n}_{c2}$, *and* $\boldsymbol{n}_{c3}$ *positively span* $\mathbb{R}^2$, *and* $\boldsymbol{n}_{c1}, \boldsymbol{n}_{c2}$, *and* $\boldsymbol{n}_{c3}$ *intersect at a single point.*

Here, the focus was on the enlarged triangle (dotted line) shown in Fig. 21. As shown in Fig. 21(b), let $l_i$ be the length of side $i$ of the enlarged triangle; $\mathrm{P}_{ci}^{l}$ be the center of the cylindrical part of Finger $i$; $l_{o2}$ be the length between $\mathrm{P}_{c2}^{l}$ and the corner between sides 1 and 2 of the enlarged triangle; $\mathrm{P}_{\mathrm{I}}^{l}$ be the intersection point described in Lemma 1; $\theta_{\mathrm{I}}$ be the angle between segments $\overline{\mathrm{P}_{c1}^{l}\mathrm{P}_{c2}^{l}}$ and $\overline{\mathrm{P}_{c2}^{l}\mathrm{P}_{\mathrm{I}}^{l}}$; and $l_r$ be the length of the side of the regular triangle formed by $\mathrm{P}_{c1}^{l}$, $\mathrm{P}_{c2}^{l}$, and $\mathrm{P}_{c3}^{l}$. Then, the conditions of Lemma 1 can be rewritten as

$$\begin{gathered} l_{o2} \sin \theta_{s12} = l_r \cos(\theta_{s12} - \theta_{\mathrm{I}}) \\ (l_2 - l_{o2}) \sin \theta_{s23} = l_r \sin(150° - \theta_{s23} - \theta_{\mathrm{I}}) \\ \sin(\theta_{s12} + \theta_{s13}) \sin(60° + \theta_I - \theta_{s12}) \\ = \sin(\theta_{s12} + \theta_{s23}) \sin(60° - \theta_{\mathrm{I}}) \end{gathered} \tag{23}$$

where, $\theta_{\mathrm{I}}$, $l_{o2}$, and $l_r$ are variables, and the remainder are the given parameters. Because the variables are uniquely determined by (23), the grasping configuration is also uniquely determined (i.e., the self-centering function is valid in State C), as validated by the results shown in Fig. 19. If the identified grasping configuration is not valid because the part of the edge that comes into contact with the finger does not exist in reality or the determined configuration is outside the movement range of the fingers, then grasping fails. If the size of the target object is within the graspable range of the gripper and the bottom-surface shape is similar to that of a triangle or regular polygon with $3n$ sides ( $n$ : a positive constant), the grasping configuration determined by (23) is valid. Moreover, grasping (second-order immobility) is feasible.

In State D, sides 2 and 3 have opposite normal directions ($\boldsymbol{n}_{c2} = \boldsymbol{n}_{c3}$), and the tip forces can be applied in these directions to achieve a force-closure grasp (in planar space). In this case, the force from finger 1 contributes slightly to grasping, as shown in Figs. 17(a) and 17(b). Thus, the risk of the object rotating when lifted increases, as shown in Fig. 17(a). Therefore, State A is preferable for objects with parallel surfaces if the robot achieves both States A and D. In State E, $\boldsymbol{n}_{c1}, \boldsymbol{n}_{c2}$, and $\boldsymbol{n}_{c3}$ do not positively span $\mathbb{R}^2$. Therefore, the object is not geometrically immobilized, and force closure is required for stable grasping. To achieve force closure, the nonzero internal force components of the tip forces ($\{\boldsymbol{f}_{\mathrm{cnt}_i} | \sum_{i=0}^{3} \boldsymbol{f}_{\mathrm{cnt}_i} = \boldsymbol{0}\}$) must satisfy the frictional condition. It is difficult to satisfy this condition when $\theta_{s12}$ and $\theta_{s13}$ increase. In States B and E, the grasping stability depends on the friction condition. Hence, if grasping in States A, C, or D is possible, then States B or E should be avoided.

## *B. Grasping strategy for prismatic objects*

Based on the results presented in the previous section, a strategy for grasping prismatic objects using the developed gripper is presented. Assuming that the position, posture, and shape of the bottom surface of the object are detected by the vision system, the strategy is to derive the desired gripper position and posture, $\boldsymbol{p}_{\mathrm{gri}} = [x_{\mathrm{gri}}, y_{\mathrm{gri}}, \theta_{\mathrm{gri}}]^{\mathrm{T}}$, for grasping a prism with an N-sided bottom located on a plane. Here, the focus is on the grasping strategies for states A, C, and D, which do not require friction information. If a valid grasping configuration cannot be obtained without considering the frictional condition (e.g., only states B or E are feasible), the use of a grasp-planning tool (e.g., GraspIt!) is necessary [17] to derive the grasping position and posture. However, considerable friction is required to achieve force closure in states B and E. Moreover, it is only in rare cases that States A, C, and D cannot be constructed because the main target objects are industrial parts whose bottom shape is similar to a regular polygon. Among the States A, C, and D, State C is the most preferred, followed by state A. The priority order (i.e., C, A, and D) is based on the following reasons. 1) In State C, second-order immobility is established, and the self-centering function is activated. 2) In State A, force closure occurs, but the self-centering function is not activated. 3) In State D, force closure occurs; however, the risk of rotation is high when the object is lifted, as shown in Fig. 17(a). The state is selected according to the priority order to derive the desired gripper position and suitable posture $\boldsymbol{p}_{\mathrm{gri}}$ for grasping in the selected state.

The grasping strategy is illustrated in Fig. 23. Initially, 3-tuple candidates for grasping surfaces that can construct second-order immobility or force closure in states A, C, or D are sought. Next, the 3-tuple of the grasping surface is selected among the candidates according to the priority order and derivation of the desired $\boldsymbol{p}_{\mathrm{gri}}$ for the tuple.

Among the edges of the bottom surface of the object detected using a vision sensor, three edges are selected. If two of the three edges are parallel, then the candidate grasping configurations to be considered are those of States A and D. If no tuples are parallel, the candidate grasping configurations are those of States B, C, and E. Specifically, the grasping configuration is that of State C because States B and E are not considered, as mentioned above. To detect the feasible 3-tuple of the grasping surface in States A, C, and D, the "grasping triangle" is defined. The grasping triangle is constructed using the three center positions of the three fingertips in contact with the selected edges, as shown in Fig. 24. The grasping triangle forms an equilateral triangle. However, these are not the three selected edges; instead, three virtual edges that are shifted away from the center by the radius ($r_{\mathrm{ft}}$) of the cylindrical part of the fingertip are considered. If two of the three (virtual) edges are parallel, the distance between them is measured. This helps to determine whether this distance is sufficiently small to construct a grasping triangle with two vertices on one side and another vertex on the other side within the motion range of the gripper. If the distance is sufficiently small, State A can be constructed, and the three selected edges are registered as candidates. Otherwise, it must be verified whether a grasping triangle can be constructed using each vertex on each (virtual) edge. If it can be constructed, State D is feasible, and the three selected edges are registered as candidates. If two of the three (virtual) edges are not parallel, it should be verified if $\theta_{s12} < 90°$ and $\theta_{s13} < 90°$ ; moreover, $\theta_{\mathrm{I}}$, $l_{o2}$, and $l_r$ must satisfy (23) to determine whether State C is feasible. If State C is feasible, the three selected edges are registered as candidates. Next, a tuple of three grasping surfaces among the candidates is selected according to the priority order, and the desired $\boldsymbol{p}_{\mathrm{gri}} = [x_{\mathrm{gri}}, y_{\mathrm{gri}}, \theta_{\mathrm{gri}}]^{\mathrm{T}}$ is derived for the tuple. The coordinate frame, $\Sigma_c$, for deriving $\boldsymbol{p}_{\mathrm{gri}}$ is set as follows: 1) the origin is located at the center of the field of view of the camera; 2) the $x_c$ axis corresponds to the horizontal direction of the camera image, and the right direction is the positive direction of the axis; and 3) the $y_c$ axis corresponds to the vertical direction of the image, and the upper direction is defined as the positive direction of the axis. The following steps determine $\boldsymbol{p}_{\mathrm{gri}}$ (see Fig. 25): 1) derive the desired gripper position, $[x_{\mathrm{gri}}, y_{\mathrm{gri}}]^{\mathrm{T}}$, corresponding to the centroid of the grasping triangle for the selected tuple; 2) derive the desired gripper posture, $\theta_{\mathrm{gri}}$, corresponding to the angle between the $y_c$ axis and the line passing through the centroid and vertex of the grasping triangle ($-120° \leq \theta_{\mathrm{gri}} \leq 120°$). Grasping experiments were conducted to validate the proposed grasping strategy. The experimental setup and representative results are shown in Fig. 26. The target object was randomly placed on a table within the field of view of the camera, as shown in Fig. 26(a). After deriving the desired gripper position and posture for grasping, the gripper was controlled to achieve the desired position and posture for grasping the object. The experiments were conducted three times for each object. All experiments were successful, indicating that the proposed strategy functioned (Fig. 26(b)).

***Grasping strategy for the developed gripper***

**Input: Bottom surface image of the target object captured by a camera**

Detect edges of bottom surfaces
***Loop*** $i, j, k = 1,2,\ldots,N$(: Num. of Surfaces) ***do***
  **If** $i$ and $j$th (virtual) edges are parallel
    **If** Equilateral triangle can be constructed on $i$ and $j$th edges
      Edges $i$, $j$, and $k$ are registered as "State A"
    **Else If** Equilateral triangle whose three vertices are on edges $i$, $j$, and $k$ can be constructed
      Edges $i$, $j$, and $k$ are registered as "State D"
    **End**
  **Else If** any two edges of $i$, $j$, and $k$ are not parallel
    **If** The conditions for "State C" ($\theta_{\mathrm{I}}$, $l_{o2}$, and $l_r$ must satisfy (23), $\theta_{s12} < 90°$, and $\theta_{s13} < 90°$) is satisfied with respect to the picked edges
      Edges $i$, $j$, and $k$ are registered as "State C"
    **End**
  **End**
**End**

**Output**:
$[x_{\mathrm{gri}}, y_{\mathrm{gri}}]^{\mathrm{T}}$ ←Centroid of grasping triangle with the highest priority
$\theta_{\mathrm{gri}}$ ←Posture of triangle

**Fig. 23.** Overall structure of the grasping strategy

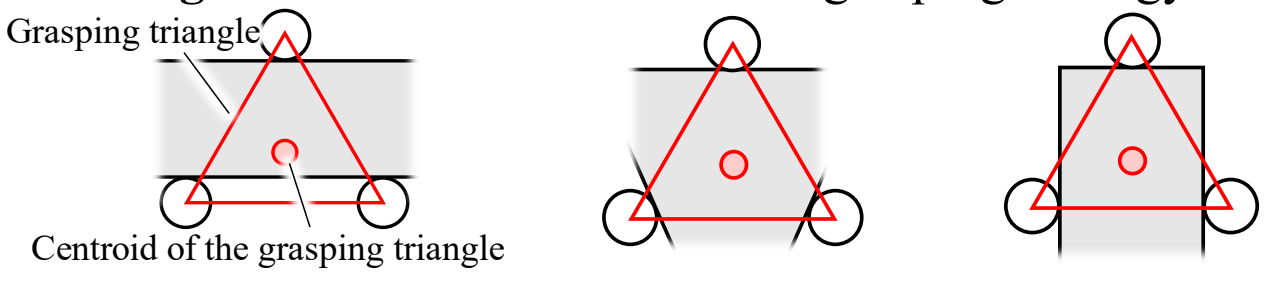

(a) State A (b) State B (c) State C

**Fig. 24.** Grasping triangle of States A, C, and D

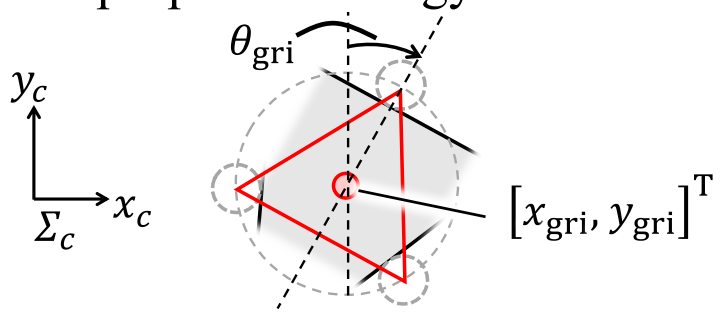

**Fig. 25.** Definition of the gripper position and posture

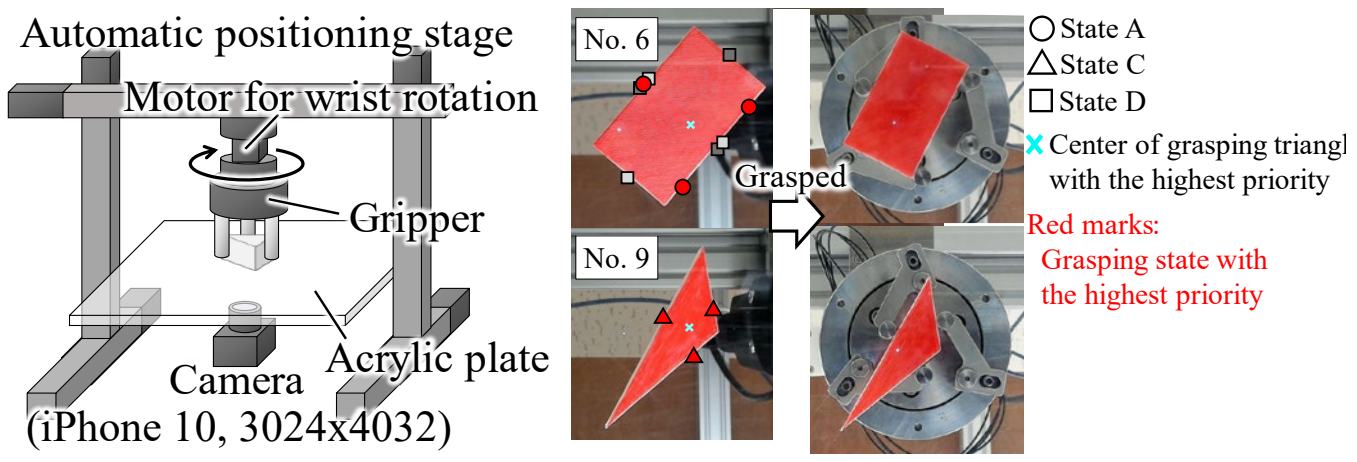

(a) Experimental setup (b) Representative results when the grasping strategy is adopted

**Fig. 26.** Evaluation of the proposed grasping strategy

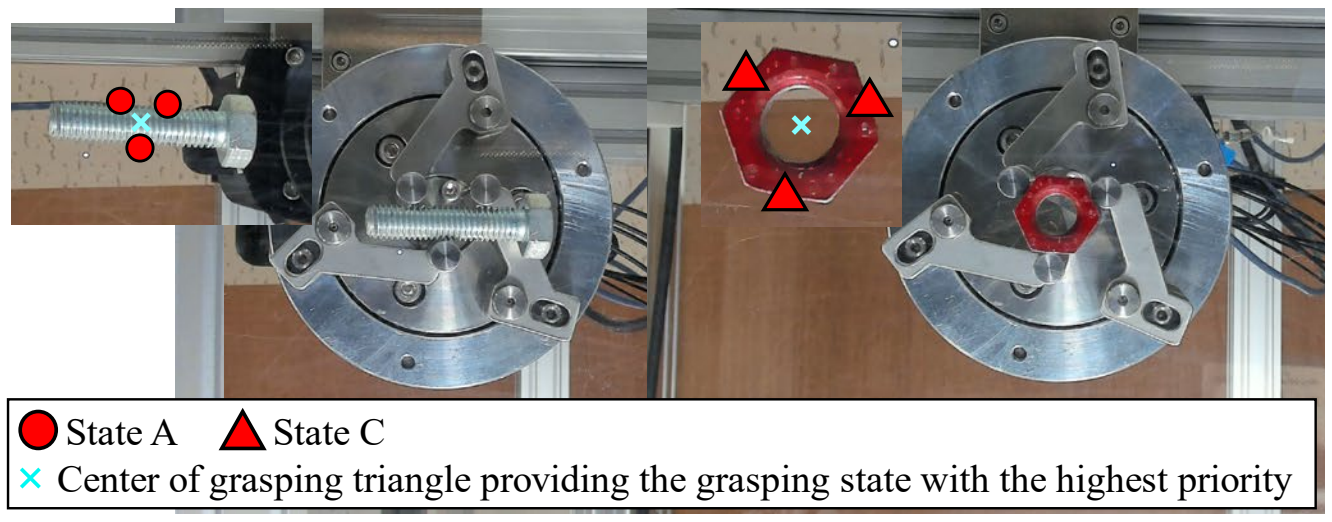

**Fig. 27.** Results of grasping test for M10 bolt and M12 nut when grasping strategy is adopted

## V. Experimental Validation

### A. Grasping test for industrial objects

The targets were industrial objects. The main targets were cylindrical objects, such as gear, pulley, and nut. The proposed grasping strategy was adopted to grasp prismatic objects. The gripper successfully grasped all objects (see the attached video). The results prove that the developed gripper can grasp small, heavy, and thin objects (e.g., gear, metal block, M3 bolt, and M8 washer). The self-centering function worked when grasping cylindrical-shaped objects. Fig. 27 shows that the proposed strategy is effective for grasping with centering. The results demonstrated the efficiency of the developed gripper.

### B. Assembly operation test

To verify the efficiency of the developed gripper during assembly, an experimental evaluation was conducted. The objective was to complete representative tasks in the industrial category challenge of the World Robot Summit in 2018 [14]. These tasks involve fundamental assembly, including picking, inserting, tightening nuts, and fixing an idler pulley. In these tasks, the target parts were initially placed on a table. The robot recognizes the position and posture of the part via the vision system and then grasps the part to perform the assembly task. As the position and posture of the part obtained by the vision-based recognition system include errors, a self-centering function is required to complete the assembly tasks. It is also necessary to reduce operation time. Therefore, this assembly operation test is suitable for evaluating the developed gripper from the perspectives of high-speed grasping motion and self-centering. In addition, the task of fixing an idler pulley is suitable for evaluating the gripper from the perspective of high force, because the gripper must fix the idler pulley while applying tension to the flexible belt. All the operational tests were successfully completed (see the attached video). The results demonstrated the efficiency of the developed gripper.

## VI. Discussion and Conclusion

In this study, a novel lightweight gripper that generates a high motion speed (maximum: approximately 1400 mm/s) and high tip force (maximum: more than 50 N throughout the graspable range and 80 N) was developed, utilizing the quick-return mechanism and LS-CVT. Table II summarizes the performance, including the closing speed, tip force, graspable range, and weight, of the developed gripper, commercial industrial grippers, and related works. The performance score $\eta$ is introduced for the comparison:

$$\eta = \frac{\text{Maximum closing speed [mm/s]}\times\text{Maximum tip force [N]}}{\text{Weight [kg]}\times 1000} \quad (24)$$

$\eta$ evaluates how much greater both closing speed and tip force can be while reducing weight. Table II shows the performance of the following grippers with known performance: three pneumatic grippers listed in [1] with the highest $\eta$ ; motor-driven grippers with high $\eta$; [3][4][5][7][8] and our previous gripper with high-speed motion [6]. As shown in this table, the developed gripper achieves the highest performance. Pneumatic-driven grippers tend to provide large $\eta$ due to their lightweight design, however the grippers require an additional power source such as an air compressor, making the entire system larger. The developed gripper achieves high performance despite using a motor-driven system. The integration of the quick-return mechanism and LS-CVT further provides the benefits of the motion range expansion for the fingertip and a reduction in the impulsive force applied to the fingers. We presented theoretical analyses of the tip force and motion speed and confirmed their feasibility through actual experiments. The results of the analyses and experiments were similar, thereby confirming that the gripper was highly effective. The developed gripper also has a self-centering function for both cylindrical and prismatic objects. This study also formulated a grasping strategy for polygonal objects. The efficiency, function, and grasping strategy of the gripper are verified using several grasping tests and assembly tasks.

Table II Performances of the Grippers

| | Driving system | Closing speed [mm/s] | Tip force [N] | Graspable range [mm] | Weight [kg] | Evaluation score $\eta$ |
|---|---|---|---|---|---|---|
| **This work** | **Motor** | **1396** | **80** | **77** | **0.3** | **372** |
| Previous work [6] | Motor | 601 | 12 | 77 | 0.2 | 36 |
| Schunk EGK 40 | Motor | 115 | 150 | 83 | 1.2* | 15 |
| Robotiq 2F-85 | Motor | 150 | 235 | 170 | 0.9 | 39 |
| Festo JPG 50 | Air | 267 | 220 | 8 | 0.2* | 288 |
| Festo HGPT 16 | Air | 273 | 106 | 6 | 0.1* | 283 |
| Sommer automatic GP404S-C | Air | 100 | 375 | 4 | 0.14* | 260 |
| [3] | Motor | N.D. | 20 | N.D. | 0.4 | - |
| [4] | Motor | N.D. | 125 | N.D. | 2.0 | - |
| [5] | Motor | 0.5 | 350 | 100 | 0.9 | 0.2 |
| [7] | Motor | N.D. | 28 | N.D. | 0.8 | - |
| [9] | Motor | N.D. | 141 | N.D. | 0.9 | - |
| [10] | Motor | N.D. | 31 | N.D. | 0.4 | - |

* Since the catalog value does not include the weight of fingers, this weight value includes the weight of the finger that is assumed to be 20 % (≈ this work) of the whole gripper weight.

## Acknowledgment

The authors would like to thank TEIKOKU CHUCK Co., Ltd. for their help with the production of the developed gripper.

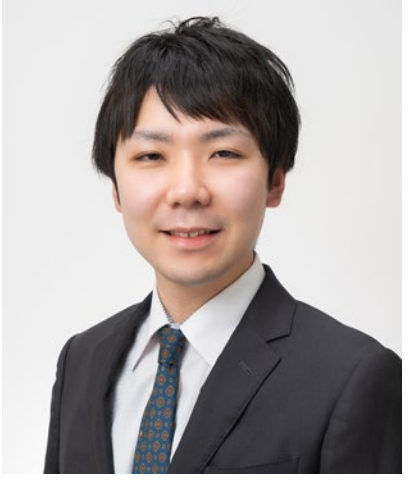

**Toshihiro Nishimura** received his BS, MS, and Doctoral degrees from Kanazawa University in 2016, 2018 and 2019, respectively. He was a researcher of industrial robots in FANUC corporation from 2018 to 2021. He is currently an assistant Professor with the Faculty of Frontier Engineering, Institute of Science and Engineering, Kanazawa University, Kanazawa, Japan. His research interests include robotic hand and grasping, object manipulation, and soft robots.

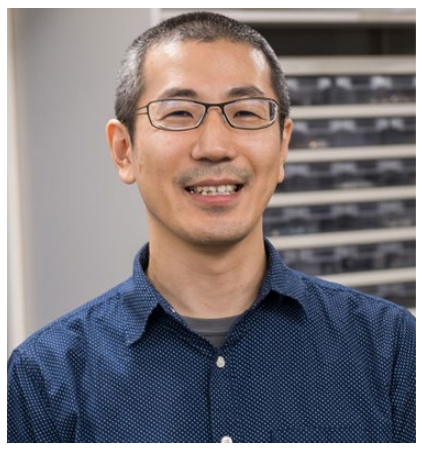

**Takeshi Takaki** received the BS and MS degrees in Mechanical Engineering from the Tokyo University of Science in 2000 and 2002, respectively, and the Ph.D. degree in Mechanical Engineering from the Tokyo Institute of Technology in 2006. He is currently a Professor with the Graduate School of Advanced Science and Engineering, Hiroshima University. His research interests include intelligent mechanisms, robot hands, continuously variable transmissions, and force visualization mechanisms.

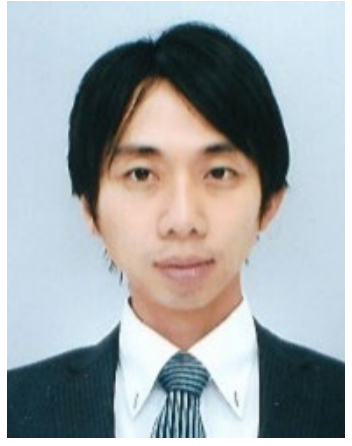

**Yosuke Suzuki** received the B.Eng., M.Eng., and Ph.D degrees in engineering from the Tokyo Institute of Technology, Tokyo, Japan, in 2005, 2007, and 2010, respectively. He is currently an assistant Professor with the Faculty of Mechanical Engineering, Institute of Science and Engineering, Kanazawa University, Kanazawa, Japan. His research interests include tactile and proximity sensors, robotic grasping, and distributed autonomous system.

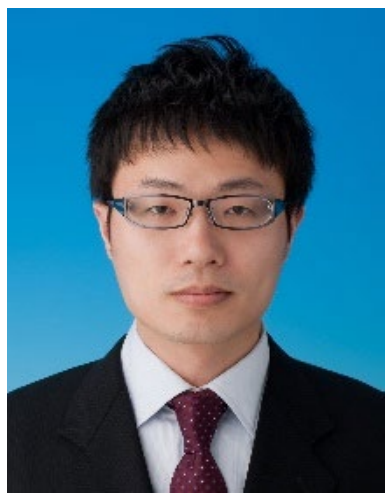

**Tokuo Tsuji** received his BS, MS, and Doctoral degrees from Kyushu University in 2000, 2002 and 2005, respectively. He worked as a Research Fellow of Graduate School of Engineering, Hiroshima University from 2005 to 2008. He worked as a Research Fellow of Intelligent Systems Research Institute of National Institute of Advanced Industrial Science and Technology (AIST) from 2008 to 2011. From 2011 to 2016, he worked as a Research Associate at Kyushu University. From 2016, he has been working as an Associate Professor at Institute of Science and Engineering, Kanazawa University. His research interest includes multifingered hand, machine vision, and software platform of robotic systems.

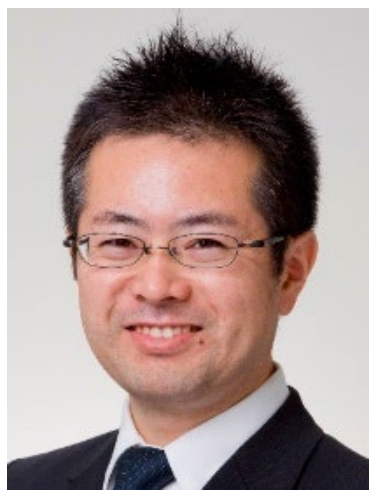

**Tetsuyou Watanabe** received the B.S., M.S., and Dr.Eng. degrees in mechanical engineering from Kyoto University, Kyoto, Japan, in 1997, 1999, and 2003, respectively. From 2003 to 2007, he was a Research Associate with the Department of mechanical Engineering, Yamaguchi University, Japan. From 2007 to 2011, he was an assistant professor with Division of Human and Mechanical Science and Engineering, Kanazawa University. From 2011 to 2018, he was an associate professor with Faculty of Mechanical Engineering, Institute of Science and Engineering, Kanazawa University. Since 2018, he has been a professor with Kanazawa University. From 2008 to 2009, he was a visiting researcher at Munich University of Technology. His current research interests include robotic hand, grasping, object manipulation, medical sensors, surgical robots, and user interface.